\pdfoutput=1
\documentclass[11pt]{article}

\usepackage{acl}
\usepackage{times}
\usepackage{latexsym}
\usepackage{textgreek}
\usepackage{mathtools}
\usepackage{amsmath}
\usepackage{txfonts}
\usepackage{xspace}
\usepackage{booktabs}
\usepackage{multirow}
\usepackage{enumitem} 
\usepackage{lipsum}
\usepackage{algorithm}        
\usepackage{subcaption}
\usepackage[noend]{algpseudocode}
\newcommand{\macrof}{$\mathrm{m}$-$\mathrm{F_1}$\xspace}
\newcommand{\microf}{$\mathrm{\muup}$-$\mathrm{F_1}$\xspace}

\usepackage{microtype}

\title{Legal-Tech Open Diaries: Lesson learned on how to develop and deploy light-weight models in the era of humongous Language Models}

\author{
  Stelios Maroudas\thanks{\hspace{0.5em}Equal contribution. Work done during capstone projects in Cognitiv+ (\url{https://www.cognitivplus.com/}).}$^{\;\;\;\dagger\diamond\;}$ \quad
  Sotiris Legkas$^{\ast\;\dagger\diamond\;}$ \\
  \bf Prodromos Malakasiotis$^{\;\dagger\;}$ \quad
  Ilias Chalkidis$^{\;\;\ddagger\diamond\;}$ \\
$^{\dagger\;}$ Department of Informatics, Athens University of Economics and Business, Greece\\
${\;\ddagger}$ Department of Computer Science, University of Copenhagen, Denmark \\
$^{\diamond\;}$ Cognitiv+, Athens, Greece\\
}

\date{}

\begin{document}
\maketitle

\begin{abstract}
In the era of billion-parameter-sized Language Models (LMs), start-ups have to follow trends and adapt their technology accordingly. Nonetheless, there are open challenges since the development and deployment of large models comes with a need for high computational resources and has economical consequences. In this work, we follow the steps of the R\&D group of a modern legal-tech start-up and present important insights on model development and deployment. We start from ground zero by pre-training multiple domain-specific multi-lingual LMs which are a better fit to contractual and regulatory text compared to the available alternatives (XLM-R). We present benchmark results of such models in a half-public half-private legal benchmark comprising 5 downstream tasks showing the impact of larger model size. Lastly, we examine the impact of a full-scale pipeline for model compression which includes: a) Parameter Pruning, b) Knowledge Distillation, and c) Quantization: The resulting models are much more efficient without sacrificing performance at large.
\end{abstract}

% 1. Introduction
\section{Introduction}

Transformer-based Languages Models (LMs) \cite{Radford2018ImprovingLU,devlin-etal-2019-bert,liu-2019-roberta} have stormed NLP benchmarks with state-of-the-art performance, while recently humongous billion-parameter-sized models \cite{brown-etal-gpt3, rae-etal-gopher, hoffman-etal-chinchilla} have showcased impressive few-shot capabilities. In addition, multi-lingual LMs \cite{conneau-etal-2020-unsupervised} have been also developed demonstrating exceptional results as well as impressive performance in zero-shot cross-lingual transfer. 

The legal NLP literature is also flourishing with the release of many new resources, including large legal corpora \cite{hendersonkrass2022pileoflaw}, benchmark datasets \cite{chalkidis2021-multieurlex, koreeda-manning-2021-contractnli-dataset, Zheng2021WhenDP,chalkidis-etal-2022-lexglue,habernal-etal-2022-argument}, and pre-trained legal-oriented language models \cite{chalkidis-etal-2020-legalbert, Zheng2021WhenDP}. Despite this impressive progress, the efficacy of differently-sized language models on legal NLP tasks and the importance of domain (legal) specificity are still understudied, while the effect of model compression techniques in model's performance and efficiency is ignored. 

In this work, we aim to shed light in all these directions following model development across three incremental steps in a \emph{pipelined} approach:\vspace{-2mm}
\begin{enumerate}[label=(\alph*),itemsep=-0.3em]
    \item \emph{model pre-training} on large legal corpora,
    \item \emph{model fine-tuning} on down-stream tasks, and
    \item \emph{model compression} to improve efficiency.
    % model's efficiency (size, speed)
\end{enumerate}\vspace{-1mm}
To do so, we initially develop 4 multi-lingual legal-oriented language models (C-XLMs). We benchmark their performance across 5 down-stream legal NLP tasks, comprising both publicly available and private datasets, covering both English and multi-lingual scenarios in several tasks types, i.e., document/sentence classification, natural language inference, and entity extraction. Finally, we experiment with a full-scale pipeline for model compression which includes a) Parameter Pruning,  b) Knowledge Distillation, and c) Quantization to produce much more efficient (smaller and faster) models that can be effectively deployed in production.

\begin{table*}[]
    \centering
    \resizebox{\textwidth}{!}{
    \begin{tabular}{ll|r|r|r|r|r|r|c|c}
         \multicolumn{2}{c}{\bf Model Alias} & \bf \#Langs &  \bf \#Layers & \bf \#Units & \bf \#Heads & \bf \#Params & \bf Vocab. Size & \bf Train. Tokens & \bf MLM Acc. \\
         \midrule
         XLM-R & base & 100 & 12 & 768 & 12 & 278M & 250k & 6.3T & 74.0\\
         XLM-R & large & 100 & 24 & 1024 & 16 & 559M & 250k & 6.3T & 78.9 \\
         \midrule
         C-XLM & tiny & 10 & 4 & 128 & 4 & 9M & 64k & 92B & 54.9 \\
         C-XLM & small & 10 & 6 & 256 & 4 & 21M & 64k & 92B & 68.9 \\
         C-XLM & base & 10 & 12 & 512 & 8 &  71M & 64k & 92B & 77.8 \\
         C-XLM & large & 10 & 24 & 1024 & 16 &  368M & 64k & 92B & 81.5\\
    \end{tabular}
    }
    \vspace{-2mm}
    \caption{Model Specifications, Training Tokens processed on pre-training and MLM performance (Acc.) for all variants of our XLM (C-XLM) models and the XLM-R models of \citet{conneau-etal-2020-unsupervised} considered as baselines.}
    \label{tab:model_specs}
    \vspace{-4mm}
\end{table*}

Our work aims to provide guidelines to legal-tech practitioners on model development (pre-training, fine-tuning, compression) bearing both performance and efficiency into consideration. Our findings show that the impact of larger vs. smaller models, domain-specific vs. generic models and the efficacy of model compression techniques varies across tasks, but in general larger domain-specific models perform better. Via full-scale model compression, we produce models with performance decrease by 2.3 p.p., while being approx. 42$\times$ smaller, and approx. 66$\times$ faster. We also find that fully compressed models outperform equally sized distilled or fine-tuned models.

% 3. Models
\section{Model Specifications}

Following \citet{chalkidis-etal-2020-legalbert}, we pre-train from scratch legal domain specific transformer-based language models. Our models are based on the RoBERTa architecture \cite{liu-2019-roberta}, i.e., trained with the Masked Language Modelling (MLM) objective, excluding the Next Sentence Prediction (NSP) one used by BERT \cite{devlin-etal-2019-bert}. 
In addition, based on the industry needs and driven by the work of \cite{conneau-etal-2020-unsupervised}, our models are a multilingual one -usually referred as XLM in the literature- and supports ten languages in total (English, French, German, Greek, Spanish, Italian, Dutch, Polish, Portuguese, Russian).

We pre-train 4 variants of custom XLM models (C-XLM) starting from a large version with 24 Transformer blocks (layers), each consisting of 1024 hidden units and 16 attention heads and continue by decreasing each time by a factor of 2 across all dimensions, i.e., blocks/layers, hidden units, and attention heads  (Table~\ref{tab:model_specs}).\footnote{A minor exception in the tiny version, where we consider 4 attention heads of 32 hidden units per head instead of 2 attention heads with 64 units per head.}

\section{Pre-Training}

\subsection{Training Corpora}
\label{sec:corpora}

We pre-trained our models using multi-lingual corpora that consist of regulations and contracts. For regulations, we used the MultiEURLEX dataset of \citet{chalkidis-etal-2021-multieurlex} that comprises 65k EU regulations officially translated in 24 languages.\footnote{In our work, we consider 9 languages (English, French, German, Greek, Spanish, Italian, Dutch, Polish, Portuguese).}. We also considered additional publicly available English resources; specifically the 250 US code books, part of the ``Pile of Law'' corpus released by \cite{hendersonkrass2022pileoflaw}, along-size 36k UK laws published by \citet{chalkidis-sogaard-2022-improved}.

Regarding contracts, we considered the LEDGAR \cite{tuggener-etal-2020-ledgar} dataset comprising 900k sections from US contracts in English; and 60k additional full contracts in English from a publicly available crawl from EDGAR. Since, there are no publicly available contracts in the rest of the languages, we translated these documents using state-of-the-art Neural Machine Translation (NMT) systems across all languages of interest.\footnote{We used the OpusMT (en2m) mBART models using the EasyNMT library.}

\subsection{Custom Vocabulary}

Relying on the above mentioned resources, we built a custom vocabulary of 64k sub-word units that better fit the documents in the respective domains and languages of interest. We opted for Byte-Pair Encodings (BPEs) \cite{sennrich-etal-2016-neural}, similarly to most recent work on Transformer-based language models~\cite{Radford2018ImprovingLU,liu-2019-roberta,conneau-etal-2020-unsupervised}.

\begin{figure*}
    \centering
    \resizebox{\textwidth}{!}{
    \includegraphics{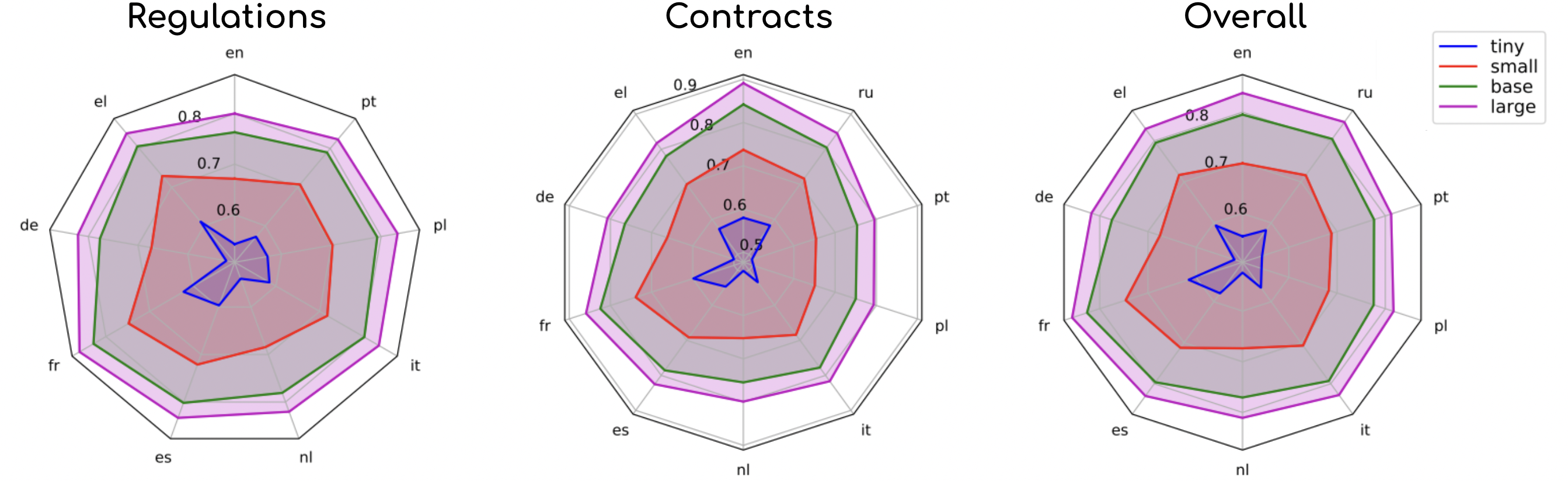}
    }
    \caption{MLM performance per language across C-XLM model variants depicted with different coloured webs.}
    \label{fig:multi_mlm}
    \vspace{-3mm}
\end{figure*}

\subsection{Masked Language Modelling (MLM)}

We pre-trained all variants of C-XLM (our domain-specific multi-lingual RoBERTa) for 1.1m steps (gradient updates) in total based on a two-step approach, similarly to \citet{devlin-etal-2019-bert}, i.e., pre-train for 1m steps with sequences up to 128 sub-word units, followed by continued pre-training for 100k steps with sequences up to 512 sub-word units, always with a batch size of 512 sequences.\footnote{This approach aims to a more efficient (compute-friendly) pre-training, since pre-training with shorter sequences severely decreases the needed compute and time.} At each example, we mask out 15\% of the tokens in total. We train all models for a maximum learning rate of $\mathrm{1e\!-\!4}$ with warm-up for the initial (5\%) training steps followed by a cosine decay.

In comparison XLM-R models were pre-trained for 1.5m steps with batches of 8192 sequences, which accounts for approx. 63$\times$ more training tokens processed; the majority of those in high-resource languages like the ones we consider.

\begin{figure}
    \centering
   \resizebox{\columnwidth}{!}{
    \includegraphics{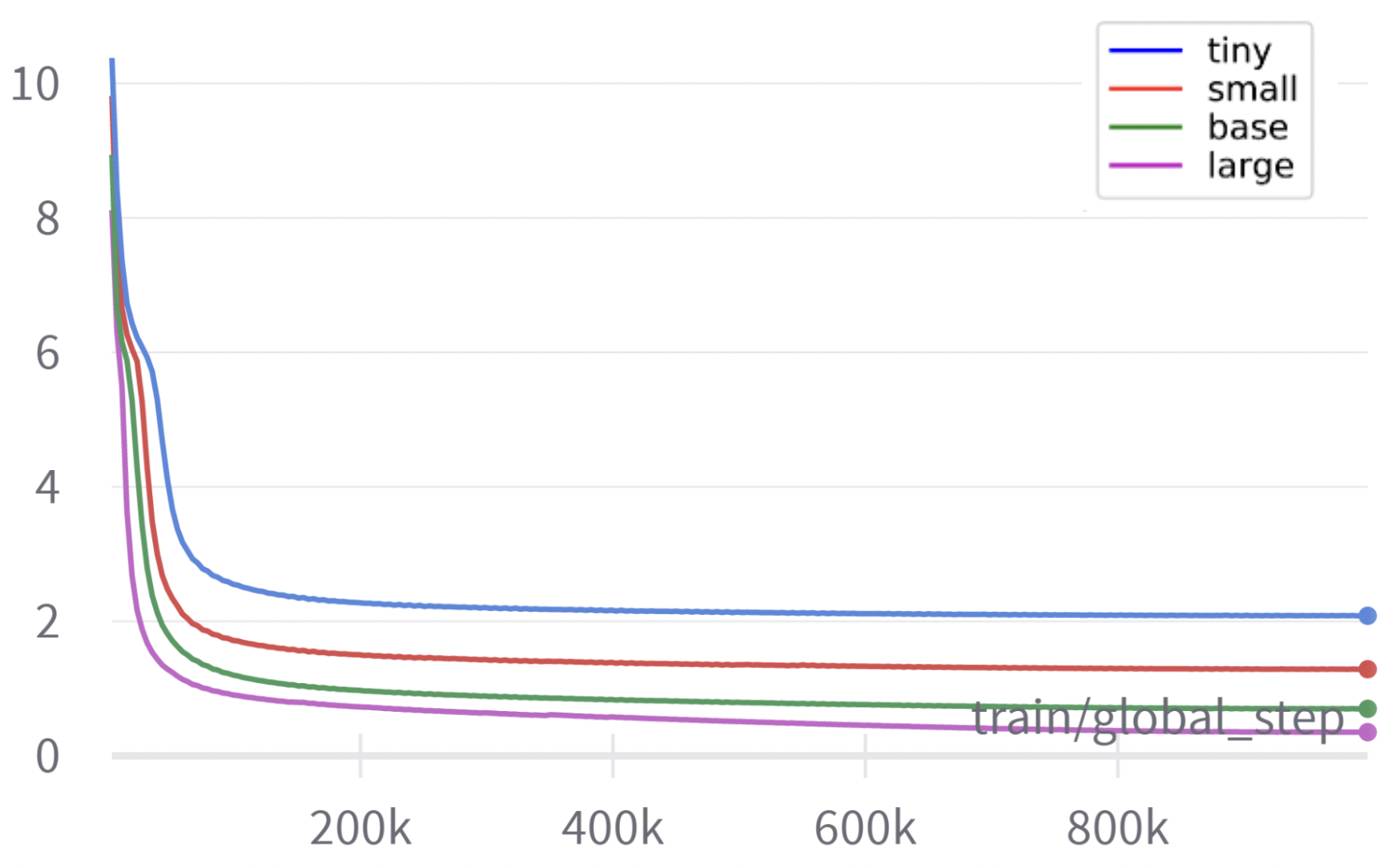}
    }
    \caption{Pre-training loss curves of C-XLMs.}
    \label{fig:loss_curves}
    \vspace{-3mm}
\end{figure}

\subsection{MLM Results}

In Figure~\ref{fig:loss_curves}, we observe the loss curves of differently sized models during pre-training. While models are equally poor performing in the very initial steps, larger models substantially outperform the smaller counterparts due to their increased capacity (number of parameters). Table~\ref{tab:model_specs} presents the accuracy of our different models. As expected, the large version (81.5\% accuracy) followed by the base version (77.8\% accuracy) of C-XLM outperform their corresponding generic XLM-R models by 2.6\% and 3.8\% respectively.\footnote{A comparison between the XLM-R models of \citet{conneau-etal-2020-unsupervised} and our models (C-XLMs) is not ideal due to the different vocabulary used. Nevertheless, it provides a general idea on pre-training performance on legal specific corpora.} Figure~\ref{fig:multi_mlm} presents masked language modelling performance in finer details across languages per model, highlighting the predominance of our two largest models.\footnote{More fine-grained MLM evaluation (per language and per document type) can be found in Appendix~\ref{sec:appendix_b}.}

%Our models (C-XLMs) consistently achieve better results in French and Greek regulations, while they achieve better results in French and English contracts (Table~\ref{tab:detailed_mlm}). 
% Finally, according to Figure~\ref{fig:multi_mlm}, the small version of C-XLM achieves better results in English contracts (+0.75\%) and is slightly outperformed by the base version of XLM-R (Table~\ref{tab:detailed_mlm}), which is impressive for its much smaller size.

\section{Fine-tuning}

\subsection{Benchmark - Tasks and Datasets}
\label{sec:benchmark}

In this section, we briefly present the evaluation benchmark that we use, which consist of both publicly available and private datasets. The benchmark is diverse covering three task types (document, sentence, and token classification) and two multi-lingual datasets.\footnote{We do not use the LexGLUE benchmark of \citet{chalkidis-etal-2022-lexglue}, since it is monolingual (English only) and also covers tasks that involve litigation, which are out of scope.} The datasets in detail are:\vspace{3mm}

\noindent\textbf{MultiEURLEX} \citep{chalkidis2021-multieurlex}, a multilingual dataset for legal topic classification comprising 65k EU laws officially translated in 23 EU languages.\footnote{MultiEURLEX is available at \url{https://huggingface.co/datasets/multi_eurlex}.} Each document (EU law) was originally annotated with relevant EUROVOC\footnote{EUROVOC is a hierarchically organized taxonomy of concepts (a hierarchy of labels) available at \url{http://eurovoc.europa.eu/}.} concepts by the Publications Office of EU. We use the 21 `Level 1' labels, obtained by \citet{chalkidis2021-multieurlex} from the original EUROVOC annotations of the documents. We use a derivative of the original dataset considering only 1k non-parallel documents per supported language (9k in total, Section~\ref{sec:corpora}).\footnote{This is inline with the work of \citet{xenuleas-etal-2022}, where the authors consider a more ``realistic'' harder version of MultiEURLEX with less and non-parallel documents.} This is a multi-label document classification task, thus we evaluate performance using macro- (\macrof) and micro- (\microf) F1 scores.\vspace{3mm}

\begin{table*}[t]
    \centering
    \resizebox{\textwidth}{!}{
    \begin{tabular}{ll|cc|cc|cc|cc|cc|cc}
         \multirow{2}{*}{\bf Model} & \multirow{2}{*}{\bf Alias} & \multicolumn{2}{c|}{\bf MultiEURLEX} & \multicolumn{2}{c|}{\bf UNFAIR-ToS} & \multicolumn{2}{c|}{\bf CNLI} & \multicolumn{2}{c|}{\bf Obligations} & \multicolumn{2}{c}{\bf ContractNER}  \\
         &  & \microf & \macrof & Acc. & MAE & \microf & \macrof & \microf & \macrof & \microf & \macrof \\
         \midrule
         XLM-R & (base) & 75.3 & 53.2 & 86.6 & 0.17 & 84.0 & 81.9 & 89.7  & 88.2 & 92.4 & 93.9 \\
         XLM-R & (large) & 77.8 & 63.8 & 89.0 & 0.16 & \bf 86.3 & \bf 84.7 & 88.9 & 87.4 & 92.8 & 93.7 \\
         \midrule
         C-XLM & (tiny) & 66.5 & 46.1 & 78.2 & 0.27 & 70.2 & 69.2  & 88.7 & 87.4 & 87.2 & 89.3 \\
         C-XLM & (small) & 72.3 & 54.7 & 85.4 & 0.20 & 79.7 & 77.0  & 90.4 & 89.0 & 90.1 & 92.4 \\
         C-XLM & (base) & 75.3 & 59.4 & 87.3 & 0.18 & 84.0 & 82.1 & 91.2 & 90.4 & 92.9 & 93.9 \\
         C-XLM & (large) & \bf 78.4 & \bf 65.4 & \bf 89.7 & \bf  0.14 & 85.3 & 83.0 & \bf 91.8 & \bf 90.6 & \bf 93.2 & \bf 94.6 \\
    \end{tabular}
    }
    \vspace{-2mm}
    \caption{Overall results of fine-tuned models across all down-stream tasks.}
    \vspace{-2mm}
    \label{tab:finetuning_results}
\end{table*}

\noindent\textbf{UNFAIR-ToS} \cite{drawzeski-etal-2021-corpus} is a dataset for detecting unfair clauses in Terms of Service (ToS) agreements from on-line platforms (e.g., YouTube, Facebook, etc.) in 4 languages (English, German, Italian, and Polish). The dataset has been annotated on the sentence-level with 8 types of \emph{unfair contractual terms}, meaning terms (sentences) that potentially violate user rights according to EU consumer law. Sentences have been also annotated according to a 3-level fairness score (\emph{fair}, \emph{partially unfair}, \emph{clearly unfair}). In our case, we examine the latter task as sentence regression and evaluate performance using Mean Absolute Error (MAE), and Accuracy (Acc.) on rounded (discrete) scores.\vspace{3mm}

\noindent\textbf{ContractNLI} \cite{koreeda-manning-2021-contractnli-dataset} is a dataset for contract-based Natural Language Inference (NLI). The dataset consists of 607 contracts, specifically Non-Disclosure Agreements (NDAs).
Each document has been paired with 17 templated \emph{hypotheses} and labeled with one out of three classes (\emph{entailment}, \emph{contradiction}, or \emph{neutral}). We examine a lenient version of this task, where instead of the full document (NDA), we represent the document with a short number of sentences which have been annotated as rationales for the specific task. This is a single-label multi-class document classification task and we evaluate performance using macro- (\macrof) and micro- (\microf) F1 scores.\vspace{3mm}

\noindent\textbf{Contract-Obligations} \cite{chalkidis-etal-2018-obligation} is a proprietary (privately developed) dataset for obligation extraction from contracts (legal agreements). The dataset consists of 100 service agreements. Each contract has been split into paragraphs (approx. 9,400 in total), and labeled with 4 obligation sub-types, i.e., \emph{Obligation}, \emph{Deliverable}, \emph{Discretion},  and \emph{Prohibition}, while some paragraphs are not relevant, resulting in a total of 5 potential classes. This is a single-label multi-class document classification task. We evaluate performance using macro- (\macrof) and micro- (\microf) F1 scores.\vspace{3mm}

\noindent\textbf{ContractNER} \cite{chalkidis-etal-2017-contracts} is a proprietary dataset for contract element extraction. The dataset consists of 3,500 contractual introductions from several types (service, employment, purchase, etc.) of contracts. Each introduction (paragraph) has been labeled with 4 entity types (\emph{Title}, \emph{Contracting Party}, \emph{Start Date},  \emph{Effective Date}). This is a single-label multi-class token classification task. Thus, we evaluate performance using macro- (\macrof) and micro- (\microf) F1 scores on entity level.

\subsection{Experimental Set Up}

We tune all models conducting a grid search for learning rates $\in$  \{1e-4, 3e-4, 1e-5, 3e-5, 5e-5, 1e-6\}. We use early stopping based on validation loss; we select and report test scores based on the model with the best validation performance.\footnote{Additional details and development scores are provided in Appendix~\ref{sec:appendix_a}}

\subsection{Fine-tuning Results}
Table~\ref{tab:finetuning_results} presents the results of the fined-tuned baselines, XLM-R models, (upper zone) and of all the variants of our C-XLM models (lower zone) for each downstream task. We hypothesize that the base and large versions of C-XLM will perform better compared to their counterpart XLM-R models. Indeed, the base version of C-XLM always outperforms XLM-R across all 5 datasets, while the large version of C-XLM outperforms XLM-R in all but one (4 out of 5) datasets.\vspace{2mm}

\noindent\textbf{MultiEURLEX:} Both large versions of C-XLM and XLM-R clearly outperform the rest of the models with the C-XLM outperforming XLM-R by 0.6 p.p. in \microf and 1.6 p.p. in \macrof. Similarly, the base version of C-XLM outperforms the equivalent version of XLM-R. Interestingly, the small version of C-XLM has comparable performance with the latter while being approx. 13$\times$ smaller.\vspace{2mm}

\noindent\textbf{UNFAIR-ToS:} Both large and base versions of C-XLM outperform their counterpart XLM-R models by 0.7 p.p. in accuracy. Again, the small version of C-XLM achieves competitive performance to base-sized models.\vspace{2mm}

\noindent\textbf{ContractNLI:} 
In this task, we find that the large version of XLM-R outperforms the one of C-XLM (+1 p.p. in \microf and +1.7 p.p. in \macrof) while both base models perform comparably. We also note that the relative differences between differently sized models are the more intense across all tasks.\vspace{3mm}

\noindent\textbf{Contract-Obligations:} On this task, all C-XLM models except the tiny version outperform the baselines (XLM-R). Specifically, the large version of C-XLM achieves +2.9 p.p. in \microf and +3.2 p.p. in \macrof compared to the large version of XLM-R.\vspace{2mm}

\noindent\textbf{ContractNER:} Similarly, our C-XLM models outperform the corresponding large and base baselines by approx. 0.5 p.p. in \microf. In addition, \macrof is higher in our large model by 0.9 p.p., while base models have identical results. Again, the small version of C-XLM is competitive to the baseline.\vspace{2mm}

\begin{figure}[t]
\centering
\begin{subfigure}[b]{\columnwidth}
\centering
   \includegraphics[width=0.85\linewidth]{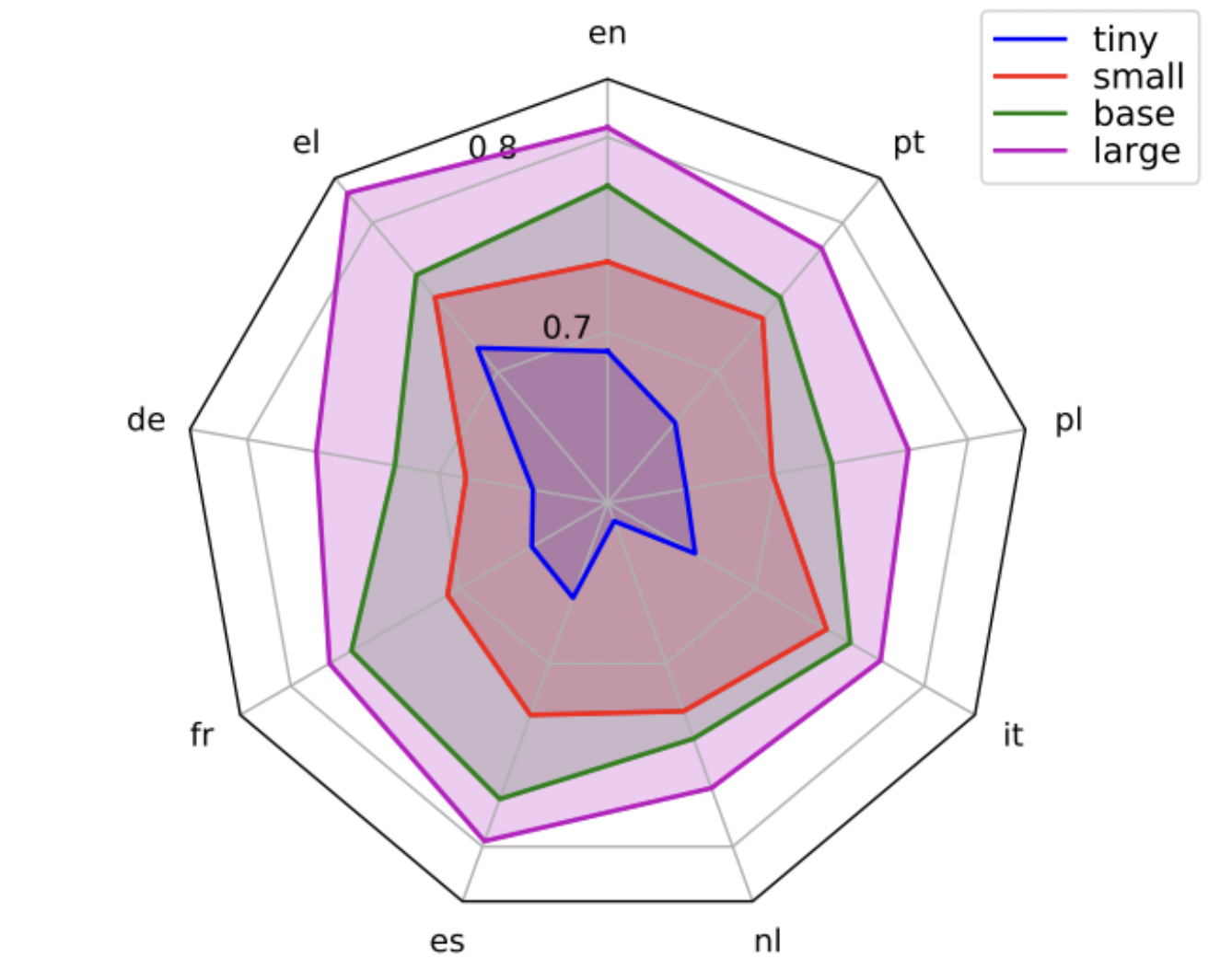}
   \caption{MultiEURLEX}
   \label{fig:Ng1} 
\end{subfigure}
\begin{subfigure}[b]{\columnwidth}
\centering
   \includegraphics[width=0.85\linewidth]{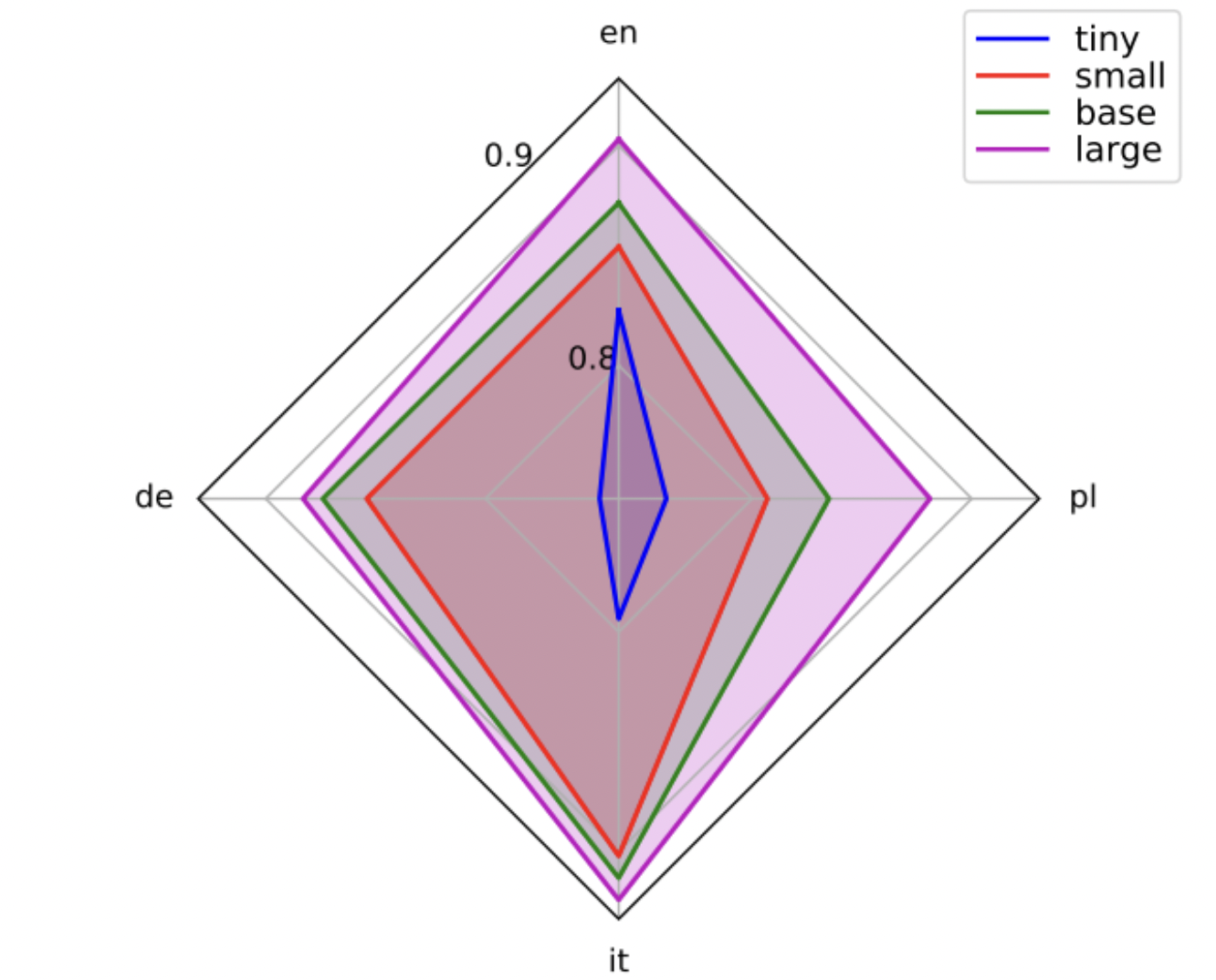}
   \caption{UNFAIR-ToS}
   \label{fig:Ng2}
\end{subfigure}
\vspace{-7mm}
\caption{Radar plots with per language performance for the multilingual MultiEURLEX and Unfair-ToS datasets for all the versions of C-XLM.}
\label{fig:multilingual_finetuning}
\vspace{-6mm}
\end{figure}

\noindent In general trends, we observe that larger models outperform smaller ones in most cases, and domain-specific models outperform generic ones, while using a sunstantially smaller (4$\times$) vocabulary and be significantly less (63$\times$) pre-trained. The largest relative differences occur in MULTIEURLEX, a 20-class multi-label classificationtask, and CNLI, a sentence pair classification task. \vspace{2mm}

\noindent\textbf{Language Parity:} Figure~\ref{fig:multilingual_finetuning} provides information through radar plots, about scores per language for each variant of C-XLM. We generally observe that performance varies across languages (e.g., models perform better in English compared to German), while also language performance disparity varies across models (depicted as differently shaped webs), and across datasets (e.g., models are better in English compared to Italian in MultiEURLEX, but the opposite is true for UNFAIR-ToS).\footnote{Refer to Appendix~\ref{sec:appendix_b} for detailed results.}

We cross out representation disparity as a possible explanation, since training data equally represent all languages (equal number of training examples). Interestingly, pre-training (MLM) accuracy also does not correlate with the down-stream performance. Based on the aforementioned points, we can only hypothesize that other qualitative characteristics (idiosyncrasies of a language in a specific context/domain) are responsible for perfomance disparities in-between languages.

\begin{algorithm}[h]
  \caption{Gradual Compression}
  \label{alg:alg1}
  \begin{algorithmic}
   \If{Teacher Size $>>$ Student Size}
  \State \textbf{S0:} Distill \emph{model} to \emph{teacher assistant} \EndIf
    \State \textbf{S1:} Prune \emph{model vocabulary} and \\ fine-tune for 1-3 epochs (if needed).
    % \If{teacher is much larger than the student}
    %     \State Distill and use teacher assistants.
    % \EndIf
     \State \textbf{S2:} Prune \emph{model depth} and distill.
     \State \textbf{S3:} Prune \emph{model width} and re-distill.
     \State \textbf{S4.1:} Optimize computational graph.
     \State \textbf{S4.2:} Apply 8-bit dynamic quantization. 
  \end{algorithmic}
\end{algorithm}
% \vspace{-4mm}

\vspace{-4mm}
\section{Model Compression}
\subsection{Methodology}
\label{sec:compression_methodology}
To compress and accelerate the inference of fine-tuned transformer-based models we adopt \emph{gradual compression}, a pipeline that combines structured pruning, knowledge distillation, and post-training quantization to progressively reach the desired compression rate, summarized in Algorithm~\ref{alg:alg1}.\footnote{See additional details and results from preliminary experiments in Appendix~\ref{sec:appendix_b}.}\vspace{2mm}

\noindent\textbf{Step 0 --- Teacher Assistant:} In case the teacher is very large and the desired compression rate is high (e.g., reducing the large version of C-XLM to the tiny one), teacher assistants \cite{Mirzadeh_2020} are used to make the transition smoother.\vspace{2mm}

\noindent\textbf{Step 1 --- Vocabulary Pruning:} The first step is to reduce the model’s vocabulary. Tokens that do not appear in the training dataset of the down-stream task are removed. 
Furthermore, using information from the tokenizer’s merges, the merge of two tokens that exist in the training dataset and individual tokens that form a merge that, also, exists in the training dataset are kept as well. 
After the redundant tokens are removed, the embedding matrix is reshaped and the new model, if necessary, is fine-tuned for 1-3 epochs, to restore its original performance. The intuition behind vocabulary reduction is that some word embeddings that were learned during pre-training might not be useful for a specific down-stream task, since such words are rare and their word embeddings would not get updated during fine-tuning, if they did not exist in the training set 
(e.g., some words of a multilingual model would be redundant for a monolingual task).\vspace{2mm}

\noindent\textbf{Step 2 --- Depth Pruning:} The second step is to reduce the model’s \emph{depth} via knowledge distillation. Similarly to \citet{sun2019patient}, we find that using the weights of the first $k$ layers from the teacher’s original pre-trained (not fine-tuned) language model produces the most consistent results.
In our implementation, the KL-divergence between the (softened) teacher’s and student’s predicted probabilities is chosen as the distillation loss function. Across all  tasks, the distillation loss is, also, linearly combined with the original loss.  For the multi-label classification task, the cross-entropy loss is replaced by a binary cross-entropy loss, again with the (softened) teacher’s and student’s probabilities as inputs, whereas for the regression task it is replaced by the mean squared error between the teacher’s and student’s output logits \cite{ba2014deep}. 
\vspace{2mm}

\begin{table*}[t]
    \centering
    \resizebox{\textwidth}{!}{
    \begin{tabular}{l|cc|cc|cc|cc|cc}
         \multirow{2}{*}{\bf Model} & \multicolumn{2}{c|}{\bf MultiEURLEX} & \multicolumn{2}{c|}{\bf UNFAIR-ToS} & \multicolumn{2}{c|}{\bf CNLI} & \multicolumn{2}{c|}{\bf Obligations} & \multicolumn{2}{c}{\bf ContractNER}  \\
         & \microf & \macrof & Acc. & MAE & \microf & \macrof & \microf & \macrof & \microf & \macrof \\
         \midrule
         \multicolumn{11}{c}{Top Bound - Performance ``Ceiling''} \\
         \midrule
         C-XLM (large) & 78.4 & 65.4 & 89.7 & 0.14 & 85.3 & 83.0 & 91.8 & 90.6 & 93.2 & 94.6 \\
         \midrule
         \multicolumn{11}{c}{Gradual Compression --- Reference C-XLM (small)} \\
        \midrule
         C-XLM (small) (FT) & 72.3 & 54.7 & \bf 85.4 & \bf 0.20 & 79.7 & 77.0 & 90.4 & 89.0 & 90.1 & 92.4  \\
         C-XLM (small) (KD) & 73.3 & 54.7 & 81.1 & 0.25 & 80.2 & 78.1 & 90.1 & 89.1 & 91.0 & 93.1\\
         \midrule
         C-XLM (large) (GC)  & \bf 74.2 & \bf 60.4 & 83.7 & 0.21 & \bf 84.5 & \bf 83.1 & \bf 92.2 & \bf 91.3 & \bf 92.2 & \bf 93.3 \\
         \midrule
         \multicolumn{11}{c}{Gradual Compression --- Reference C-XLM (tiny)} \\
         \midrule
         C-XLM (tiny) (FT) & 66.5 & 46.1 & 78.2 & 0.27 & 70.2 & 69.2 & 88.7 & 87.4 & 87.2 & 89.3 \\
         C-XLM (tiny) (KD) & 64.0 & 42.0 & 76.7 & 0.30 & 75.3 & 74.3 & 89.1 & 88.1 & \bf 87.7 & 90.1 \\
         \midrule
         C-XLM (large) (GC) & \bf 73.2 & \bf 57.0 & \bf 79.6 & \bf 0.25 & \bf 80.7 & \bf 79.2 & \bf 91.9 & \bf 90.7 & 87.6 & \bf 90.2 \\
         \bottomrule
    \end{tabular}
    }
    % \vspace{-2mm}
    \caption{Model compression results across down-stream tasks. We report the performance for two baselines: (a) fine-tuning the reference pre-trained C-XLM model (FT), and (b) Knowledge Distillation and Vocabulary Pruning. where the student is the reference pre-trained C-XLM (KD); alongside the performance of fully gradually compressed (GC) models, i.e., pruned, distilled and quantized (P+KD+Q). We report the model's performance across the incremental compression steps (S) presented in Section~\ref{sec:compression_methodology} in the Appendix (Table~\ref{tab:compression_perfomance_steps}).}
    \label{tab:compression_perfomance}
    \vspace{-4mm}
\end{table*}

\noindent\textbf{Step 3 --- Width Pruning:} Once the fine-tuned teacher’s knowledge is distilled to the student model, structured pruning is applied to reduce the student's \emph{width}. In particular, using {\it TextPruner} \cite{yang-etal-2022-textpruner}, the top $n$ neurons from the intermediate fully-connected layers and the top $a$ attention heads from the multi-head attention layers that have the smallest impact on the expected loss are iteratively removed \cite{michel2019sixteen,prasanna2020bert}. The pruned student model is re-distilled to restore its original performance. Although unstructured pruning \cite{han2015learning, sanh2020movement, louizos2017learning} would probably lead to higher compression rates with smaller performance loss, we choose structured pruning to ensure that the compressed model’s inference speed is also accelerated.\vspace{2mm}

\noindent\textbf{Step 4 --- Graph Optimization \& Model Quantization:} For the final step, the student’s weights are quantized to 8-bits, using post-training dynamic quantization. However, although 8-bit quantization will reduce the memory footprint by approximately 4x, without specialized hardware there will be hardly any inference time speed-up. Thus, before quantizing the student model, using {\it ONNX} \cite{bai2019}, its computational graph is optimized, which can provide hardware-independent acceleration \cite{Li_2021}. 
In particular, constant folding --where constant expressions are statically pre-computed--, redundant node elimination --where redundant nodes such as identities are removed without changing the graph structure-- and operation fusion --where multiple smaller nodes are fused into one, reducing in this way launch and synchronization overhead \cite{Vasilache_2018}-- 
are applied.\\

\noindent\textbf{Why gradual compression?} Although gradual compression can be more time-consuming than, for example, distilling the teacher’s knowledge in a student with a smaller predefined size, it offers more flexibility and control over the whole compression process. When the desired compression rate is reached gradually, one could better balance the performance/compression-rate trade-off. 

If for example the model is sensitive to reducing the depth, one could prune the width more aggressively and vice versa. Since the model will only be compressed once before deployed, it is important to ensure that the productionized model will perform as well as possible, thus, devoting more time to take careful steps should not be a concern. 

\subsection{Compression Results}
For each down-stream task, the goal is to produce compressed versions of the large and base C-XLM that can outperform the fine-tuned small and tiny variants of C-XLM, while being smaller and faster in terms of memory and inference speed. Using Gradual Compression (GC), the final compressed versions with the small version of C-XLM as a reference comprise 6 Transformer blocks, 24 attention heads, and 1024 units, whereas the compressed versions  with the tiny one as a reference consist of 3 blocks, 12 heads and 512 units. 

The large C-LXM used as the teacher is substantially larger (20-40$\times$) compared to the reference models. To ensure that the transition to compressed versions is smooth, we first distill it using as student the base version of C-XLM, to create a \emph{teacher assistant} \cite{Mirzadeh_2020}. In every incremental step of GC where knowledge distillation is applied, the learning rate, temperature and $a$ (the original and distillation loss weighing) are tuned using grid search. Our GC compression pipeline is also compared with a variant of {\it Pre-trained Distillation} \cite{turc2019}, where the teacher’s (or its assistant’s) knowledge is distilled directly to the reference (smaller) pre-trained model.

Results are presented in Table~\ref{tab:compression_perfomance}. We observe that the compressed versions of the large C-XLM model (GC) produced by the full-scale compression pipeline introduced in Section~\ref{sec:compression_methodology} always outperform both the respective fine-tuned (FT) models of the smaller versions of C-XLM and the distilled (KD) ones with a single exception in UNFAIR-ToS. Similar results can be derived when we consider the base version of C-XLM as the teacher model. Results are presented in Appendix~\ref{sec:appendix_b}.
% A possible explanation could be that since it is a regression task, knowledge distillation cannot achieve its full potential, as due to the lack of softmax in the classification head, softened probabilities among the classes are not produced, and only the output logits are used as the student’s target . However, in the tiny-models setting, our method can produce a model that is able to outperform the original fine-tuned model by 1.4 micro-F1 percentage units.

The largest relative differences are observed in the setting where we use the tiny model as a reference, which indicates that gradual compression is very effective when higher compression rates are being considered. Interestingly, in the Obligation extraction task, the compressed models are able to outperform the teacher (Upper Bound).
% Our compression pipeline also outperforms {\it Pre-trained Distillation} in every task and setting, except for the {\emph ContractNER} task in the tiny-models setting, where it produces a micro-F1 score that is smaller by 0.1 percentage units (but a larger macro-F1 score). 

% It is important to note that in the setting where the tiny version of C-XLM is the reference model, the gradually compressed models have one less transformer block (although larger hidden size) compared to the tiny version of C-XLM, but still outperform in all but one case. 

Results can be  vastly improved if a more fine-grained network-architecture search is conducted. For example, in some of the tasks, the largest performance drop occurs during the second step (depth pruning). This could be prevented if few additional layers remain, in favor of aggressive width pruning (step 3).\footnote{The size and inference speed of models produced at each compression step are reported in Appendix ~\ref{app:additional_experiments}.} However, the goal of our experiments was to produce competitive results among all tasks, even with the constraint of using shared predefined network specifications.

\begin{table}[t]
    \centering
    \resizebox{\columnwidth}{!}{
    \begin{tabular}{l|c|c|c}
         \multirow{2}{*}{\bf Model} & \multicolumn{1}{c|}{\bf Performance} & \multicolumn{1}{c|}{\bf Compression} & \multicolumn{1}{c}{\bf Inference} \\
         & \bf Loss & \bf Rate & \bf Acceleration \\
         \midrule
         \multicolumn{4}{c}{Reference C-XLM (small)} \\
         \midrule
         FT & -4.1 p.p. & 17.4$\times$ & 34.1$\times$ \\
         KD & -4.5 p.p. & 21.0$\times$ & 36.2$\times$ \\
         \midrule
         GC & \bf -2.3 p.p. & \bf 41.8$\times$ & \bf 65.5$\times$ \\
         \midrule
         \multicolumn{4}{c}{Reference C-XLM (tiny)} \\
         \midrule
         FT & -9.5 p.p. & 40.3$\times$ & 87.9$\times$ \\
         KD & -9.1 p.p. & \bf 50.9$\times$ & 94.2$\times$ \\
         \midrule
         GC & \bf -5.1 p.p. & \bf 50.9$\times$ & \bf 169.8$\times$ \\
    \end{tabular}
    }
    \vspace{-2mm}
    \caption{Averaged performance and efficiency statistics for each model across all tasks.}
    \label{tab:avg_compression_stats}
    \vspace{-4mm}
\end{table}

\begin{figure}[t]
    \centering
    \includegraphics[width=\columnwidth]{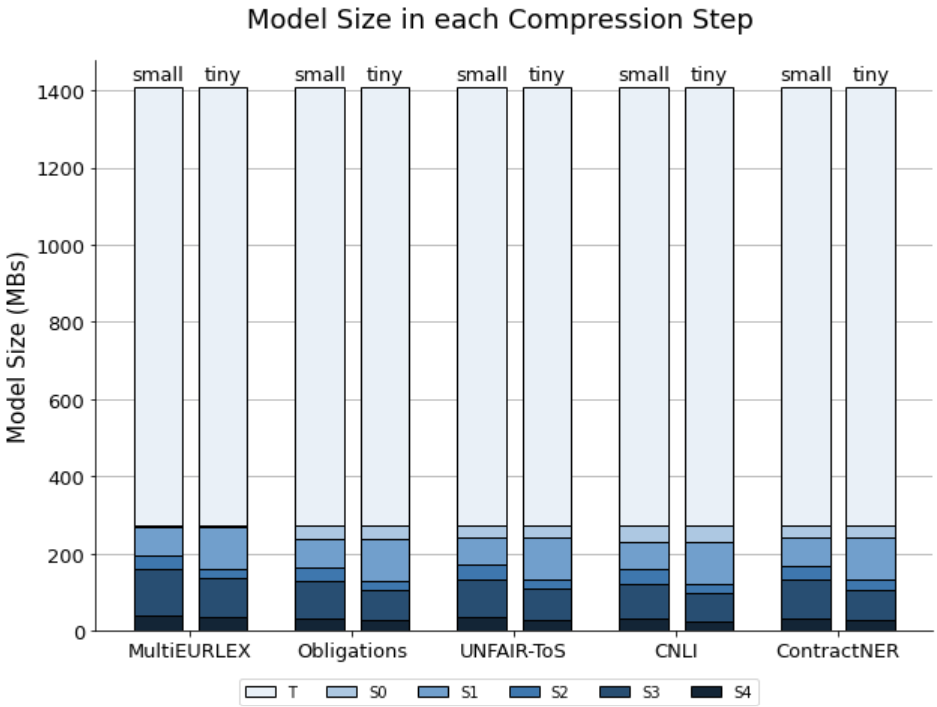}
        % \vspace{-5mm}
    \caption{Model size (MB) in each compression step (S) in relevance to the original model, C-XLM (large).}
    \label{fig:compression_sizesl}
    \vspace{-4mm}
\end{figure}

\subsection{Efficiency Considerations}
\label{sec:efficiency_considerations}
% To determine the inference speed-up of each model, we use the average number of seconds to predict a 32-batch on CPU (Ryzen 7 4700u). 
In Table~\ref{tab:avg_compression_stats}, we present aggregated (averaged) statistics in terms of efficiency.\footnotemark[14]
With the small version of C-XLM as a reference, GC produces models that are 41.8$\times$ smaller and 65.5$\times$ faster, while losing only 2.3 p.p. of performance on average. On the other hand, the fine-tuned (FT) or distilled (KD) models have a larger performance drop (by approx. 1-2 p.p.) compared to the GC versions which are also substantially (almost 2$\times$) faster on average.

With the tiny version of C-XLM as a reference, GC can produce models that are, on average, 50.9x smaller and 169.8x faster, while losing 5.1 p.p. of performance on average. The fine-tuned (FT) or distilled (KD) models have now substantially larger performance drop (9 p.p.) highlighting the benefits of GC in an extreme-compression setting. 

In Figure~\ref{fig:compression_sizesl}, we present the model size reduction across the incremental GC steps (S0-S4). The largest size reduction in both settings (small, tiny) is observed in the quantization step (S4), if we exclude the preliminary distillation step to create the teacher's assistant (S0), which reduces the original model size approx. 4$\times$.\footnote{The size compression effect of steps S1-S4 is better depicted in Figure~\ref{fig:compression_sizesb} in Appendix~\ref{sec:appendix_b} where the base version of C-XLM is the teacher and thus S0 is omitted.}

% although C-XLM (KD) and GC compression rates are the same, GC models are much faster, demonstrating its efficacy over the inference acceleration aspect (results for each compression step can be found in table~\ref{tab:compression_size_speed} of Appendix~\ref{sec:appendix_b}).

% In particular, GC can produce models that are, on average, 50.9x smaller and 169.8x faster, while loosing 5.1\% of performance on average (Table~\ref{tab:avg_compression_stats}). When compared to C-XLM (KD) and C-XLM(FT) Tiny, their performance drop is much larger (more than 9 percentage units), actively validating the benefits of GC in an extreme-compression setting. 

% 2. Related work
\section{Related Work}

\subsection{Transformer-based LMs}

\citet{devlin-etal-2019-bert} are the first to pre-train transformer-based language models (BERT) on large corpora that achieving state-of-the-art results in generic NLP benchmarks (\cite{wang-2019-glue, wang-2019-superglue}. One year later, \citet{liu-2019-roberta} argued that BERT was significantly under-trained and introduced RoBERTa (Robustly optimized BERT) using improved pre-training settings (more data, larger mini-batches, dynamic masking, and a larger vocabulary) leading to new state-of-the-art results.

Moreover, multilingually pre-trained models \cite{NEURIPS2019_c04c19c2,conneau-etal-2020-unsupervised} have been developed using a shared vocabulary, which can later fine-tuned across several languages. These models have also shown to have exceptional  zero-shot cross-lingual capabilities, a direction that we do not investigate in this work. 

In the NLP literature, domain-specific models outperform generic ones in domain-specific benchmarking. \citet{Lee_2019} created BioBERT by further pre-training BERT of \cite{devlin-etal-2019-bert} on biomedical corpora. 
n the same manner, \citet{alsentzer-etal-2019-publicly} further pre-trained BioBERT on clinical notes, releasing ClinicalBERT. 
Similarly, \citet{beltagy-etal-2019-scibert} pretrained BERT model on scientific publications called SciBERT, while \citet{loukas-etal-2022-finer} released SEC-BERT pre-trained on US financial filings. 

In the legal domain, \citet{chalkidis-etal-2020-legalbert} released LegalBERT, a legal-oriented BERT model pre-trained on diverse English legal corpora, which outperform generic ones in most legal NLU benchmark \cite{chalkidis-etal-2022-lexglue} as is CaseLawBERT of \citet{Zheng2021WhenDP}, a BERT model pre-trained solely on US case law. Recently, \cite{hendersonkrass2022pileoflaw} released a new legal-oriented larger BERT model, which is also heavily biased towards legal proceedings in US-based jurisdictions.

\subsection{Model Compression}

Unstructured pruning
% , that is, setting to zero unimportant connections, 
was popularized by 
% Han et al. (\citeyear{han2015learning}), 
\citet{han2015learning}, who iteratively located and pruned connections whose weights were less than a pre-specified threshold and retrained the sparsed network. In later work, the idea of learning how to sparsify models during training was also proposed \cite{zhu2017prune, louizos2017learning, sanh2020movement}. For Transformer-base models, \citet{sanh2020movement} argued that changes in weights during fine-tuning must be taken into consideration and proposed {\it movement pruning}.
% , a strategy that moves the selection criteria from the 0th to the 1st order. 
% On the other hand, structured pruning produces smaller (not sparse) 
% % and thus accelerated networks, 
% by removing whole units from neural networks. For multi-head attention layers, \citet{voita2019analyzing} pruned attention heads by using a differentiable 
% % relaxation of the 
% $L_0$ regularizer proposed by \citet{louizos2017learning}. \citet{mccarley2019structured}) also extended the work of \citeauthor{louizos2017learning} to attention heads and neurons from fully-connected layers. In the same direction, \citet{michel2019sixteen} proposed an iterative pruning technique, where they 
% % iteratively 
% removed heads, using the expected sensitivity of the loss to the output of each attention head. \citet{prasanna2020bert} extended the work of \citeauthor{michel2019sixteen} 
% % to also work 
% on neurons of fully-connected layers. Lastly, \citet{lagunas2021block} introduced a block pruning technique, by extending movement pruning of \citet{sanh2020movement} to work on weight matrices partitioned into fixed-sized blocks, rather than individual weights.

On the other hand, structured pruning produces smaller (not sparse) models by removing attention heads \cite{voita2019analyzing, mccarley2019structured, michel2019sixteen}), individual \cite{mccarley2019structured, prasanna2020bert} or blocks \cite{lagunas2021block} of neurons from fully-connected layers in a structured manner. We follow this line of work, since structured pruning improves model compression in practice (deployment of smaller models), contrary to unstructured pruning which sparsify networks (deployment of sparse, but equally-sized models comparing to the original ones).

% Another approach to reduce the memory footprint of neural networks is quantization. With quantization, a set of continuous real-valued numbers (that is, parameters and activations in a deep learning setting) are mapped over a fixed set of discrete numbers to minimize the number of bits required to store them. When transformer-based models are quantized to 8-bits, the models’ memory overhead is reduced approximately by 4x \cite{bondarenko2021understanding}, while the matrix multiplication computational cost can be reduced by 3.7x with the use of specialized hardware and VNNI instructions \cite{bhandare2019efficient}. \citet{junczys2018marian} and \citet{bhandare2019efficient}) applied 8-bit post training quantization to transformer-based models for machine translation, where the latter also demonstrated how to utilize specialized hardware for faster inference. \citet{zafrir2019q8bert}) used quantization aware training to quantize the embedding and the fully-connected layers of a BERT model and found that results were on par (or even exceeded) the results full-precision ones on GLUE and SQuAD tasks. Lastly, \citet{fan2020training}) added quantization noise (that is, mimicking quantization during the forward pass) only to a random subset of parameters during training and compressed a RoBERTa model by 34x, while maintaining close to full precision performance.

Another approach to reduce the memory footprint of neural networks is quantization, i.e., mapping the real-valued parameters and activations over a fixed set of discrete numbers to minimize the number of bits required to store them. When transformer-based models are quantized to 8-bits, the models’ memory overhead is reduced approximately by 4x \cite{bondarenko2021understanding}, while the matrix multiplication computational cost can be reduced by 3.7x with the use of specialized hardware.
% and VNNI instructions \cite{bhandare2019efficient}. 
\citet{junczys2018marian} and \citet{bhandare2019efficient}) applied 8-bit post training quantization to transformer-based models. 
\citet{zafrir2019q8bert} and \citet{fan2020training} used quantization aware training to quantize transformer-based language models. 
% Lastly, \citet{fan2020training}) added quantization noise only to a random subset of parameters during training and compressed a RoBERTa model by 34x, while maintaining close to full precision performance.

The last technique that is frequently used to compress transformer-based models is knowledge distillation. With knowledge distillation, a smaller ({\it student}) network is trained to mimic the behavior of a larger ({\it teacher}) network. In particular, instead of training the student network with the true labels, the teacher’s predictions are used as a target ({\it response-based knowledge distillation}), which are usually “softened” to better capture similarities across classes \cite{hinton2015distilling}. 

Along with the teacher’s predictions, information from the teacher’s intermediate states ({\it feature-based knowledge distillation}) such as hidden states \cite{sun2019patient}, embeddings \cite{jiao2019tinybert} and attention distributions \cite{sun2020mobilebert} have been used, an interesting direction that we do not explore in this work. 
% However, punishing the student whenever it deviates from the teacher’s features can be rather restrictive, thus information from the teacher’s feature-relations ({\it relation-based knowledge distillation}) has also been proposed, such relations between values \cite{wang2020minilm}, queries and keys \cite{wang2020minilmv2}, to enhance the student’s performance. 

\section{Conclusions}

Following model development across all three incremental steps of the examined \emph{pipelined} approach, we make the following observations:
\begin{enumerate}[label=(\alph*),itemsep=0em]
    \item Larger models outperform smaller ones; the performance increase varies across tasks.
    \item Domain-specific models outperform generic ones, although gains are decreased considering much large models.
    \item  Fully compressed (pruned, distilled, and quantized) models severely outperform equally sized distilled or fine-tuned models. 
\end{enumerate}

To conclude, our guidelines to LegalTech practitioners who aim to build effective, but also efficient models, can be summarized in four general points: 
\begin{enumerate}[itemsep=0em]
    \item Pre-train large-scale domain-specific language models, if possible; in case there are no such models already available.
    \item  Fine-tune the largest possible model available based on your compute capabilities.
    \item Compress the fine-tuned models to derive much smaller models that can efficiently be deployed in production; consider a  suitable compression rate to balance the performance / efficiency trade-offs. 
    \item Follow a full-scale compression pipeline (Vocabulary Pruning, Parameter Pruning, Knowledge Distillation, Graph Optimization and Quantization) for best results.
\end{enumerate}

\section*{Broader Impact and Ethics Considerations}
In this sections, we would like to discuss the broader impact and ethical considerations with respect to the use of data, privacy issues and environmental considerations.\vspace{2mm}

\noindent\textbf{Use of Data} In this work, we considered two sources of open publicly available data. The first source is legislation from EU \cite{chalkidis-etal-2021-multieurlex} published by the EU Publication Office,\footnote{\url{https://eur-lex.europa.eu/}} UK \cite{chalkidis-sogaard-2022-improved} published the UK National Archives,\footnote{\url{https://www.legislation.gov.uk/}} and US \cite{hendersonkrass2022pileoflaw} published by the U.S. Government Publishing Office.\footnote{\url{http://www.gpo.gov/}} The second source is  US contracts \cite{tuggener-etal-2020-ledgar, borchmann-etal-2020-contract} published as exhibits in public filings at SEC-EDGAR.\footnote{\url{https://www.sec.gov/edgar/}} As discussed in \citet{hendersonkrass2022pileoflaw}, the content from these legal sources implicitly encodes privacy and toxicity rules since its content is handled by governments and courts, contrary to generic web material scraped from the web \cite{dodge-etal-2021-documenting}. 

In another note, many of these sources that we used to pre-train our C-XLM models, overlap with the benchmark datasets we used to evaluate the very same models, e.g., the MultiEURLEX dataset used both for pre-training and evaluation (Sections~\ref{sec:corpora} and~\ref{sec:benchmark}). As \citet{krishna-etal-2022-computing} recently showed using downstream datasets make surprisingly good up-stream (pre-training) corpora, if domain specificity and such applications is the goal, in contrast to heavy generalization across domains and acquirement of common knowledge.\footnote{Of course, we always consider fair evaluation practices, i.e., no access to the test subsets of evaluation datasets.} \vspace{2mm}

\noindent\textbf{Environmental Considerations} Modern large deep learning models are cost intensive financially to train, due to the cost of hardware, electricity -especially in these challenging times-, and cloud compute \cite{strubell-etal-2019-energy}. They are also environmentally expensive due to the operational carbon footprint, i.e., carbon emissions, \cite{dodge-etal-2022-measuring}. It has been also demonstrated that the impact of deployment and inference can be equally or more harmful compared to training with regards to carbon emissions \cite{wu2022sustainable}, hence effective counter-measures should be considered to compensate for the financial and environmental cost.
    
By compressing and accelerating larger models, the carbon footprint of inference can be significantly reduced as we show in Section~\ref{sec:efficiency_considerations}; compensating in this way (on the long run) the environmental implications of large-scale training. Furthermore, by decreasing their memory requirements (model size, and architecture complexity), predictive models can be hosted on more environmentally-friendly infrastructure, e.g., moderate-compute cloud servers with low memory and processing power leading to a decreased energy footprint, contrary to high-end energy-intensive GPU-accelerated machines.\vspace{2mm}
    
\noindent\textbf{Privacy Considerations} Privacy concerns are also a critical topic, especially in the legal-tech industry, since prospect users (law firms, companies, and civilians etc.) want to process large quantities of documents, many of which include confidential information (e.g., private contracts). While there are many directions to privacy preserving ML via differential privacy \cite{abadi-etal-2016-privacy,klymenko-etal-2022-differential} or federated learning \cite{ryffel-etal-2018}, the problem of data leakage is practically unsolved, since the risks of sharing private documents are not considered and the responsibilities are transferred to data and cloud security.
    
Since highly accurate compressed models are able to be developed (Section~\ref{sec:compression_methodology}), deployed and run on moderate-compute servers (Section~\ref{sec:efficiency_considerations}), such technologies can be deployed on premises as an in-house solution on private clouds; or even run on the client side on server-client web infrastructures, eliminating the need for hosting data remotely or using API calls to remote cloud servers over the web, thus effectively contribute in a safer, more secure (private) AI.

\section*{Limitations} 

Based on our experiments, similarly to the literature, there no is free lunch with respect to model compression, and further compressing models takes a toll on performance. Experimenting with much larger models and examining their performance and potential for compression, following the line of work of \citet{rae-etal-gopher, hoffman-etal-chinchilla} would be fascinating but we lack resources to built billion-parameter-sized models, while increasing resources would have a larger impact with respect to environmental considerations. Based on the findings of \citet{hoffman-etal-chinchilla}, our models are not under-trained, and exploring larger models would have to be followed by an analogous increase of pre-training data and compute.

\section*{Acknowledgments}
This research has been co‐funded by the European Regional Development Fund of the European Union and Greek national funds through the Operational Program Competitiveness, Entrepreneurship and Innovation, under the call RESEARCH – CREATE – INNOVATE (Τ2ΕΔΚ-03849).  This work is also partly funded by the Innovation Fund Denmark (IFD)\footnote{\url{https://innovationsfonden.dk/en}} under File No.\ 0175-00011A.

This project was also supported by the TensorFlow Research Cloud (TFRC)\footnote{\url{https://sites.research.google/trc/about/}} program that provided instances of Google Cloud TPU v3-8 for free that were used to pre-train all C-XLM language models. Cognitiv+ provided the compute (16$\times$ Quadro RTX 6000 24GB) to fine-tune all models.

\bibliography{anthology,nllp2022}
\bibliographystyle{acl_natbib}
\appendix

\begin{table*}[t]
    \centering
    \resizebox{\textwidth}{!}{
    \begin{tabular}{l|cccccccccc|c}
         \multirow{2}{*}{\bf Corpus} & \multicolumn{10}{c}{\bf Tokens per Language} \\
         & EN & EL & DE & FR & ES & IT & NL & PL & PT & RU & All\\
         \midrule
         Contracts & 125.3M & 111.1M & 103.3M & 121.7M & 122.5M	& 113.4M & 110.7M & 89.8M & 111.3M & 89.6M & 1.1B\\
         Regulations & 178.7M & 71.4M & 62.8M & 74.0M & 77.1M & 70.1M & 71.2M & 31.9M & 71.5M & - & 708.7M\\
         \midrule
         All & 304M & 182.5M & 166.1M & 195.7M & 199.6M & 183.5M & 181.9M & 121.7M & 182.8M & 89.6M & 1.8B\\
         \midrule
    \end{tabular}
    }
    \caption{Total tokens used per language per pre-training corpus.}
    \label{tab:corpora_tokens}
\end{table*}

\begin{table*}[t]
    \centering
    \resizebox{\textwidth}{!}{
    \begin{tabular}{l|cc|cc|cc}
         \multirow{2}{*}{\bf Model} & \multicolumn{2}{c|}{\bf CNLI} & \multicolumn{2}{c|}{\bf Obligations} & \multicolumn{2}{c}{\bf ContractNER} \\
         & \microf & \macrof & \microf & \macrof & \microf & \macrof \\
         \midrule
         \multicolumn{7}{c}{Baselines} \\
         \midrule
         C-XLM Base (Ceiling Baseline) & 84.0 & 82.1 & 91.2 &90.4 & 92.9 & 93.8 \\
         C-XLM Tiny (Bottom Baseline) & 70.2 & 69.2 & 88.7 & 87.4 & 87.2 & 89.3 \\
         \midrule
         \multicolumn{7}{c}{S1: Vocab Pruning} \\
         \midrule
         S1.1: Prune vocabulary at random & 79.4 & 77.2 & 88.7 & 87.6 & 88.5 & 83.2 \\
         S1.2: Prune vocabulary based on training data & \bf 84.8 & \bf 82.9 & \bf 91.7 & \bf 90.6 & \bf 93.2 & \bf 94.1 \\
         \midrule
         \multicolumn{7}{c}{Input Model: S1.2  --$>$ S2: Pruning Depth (Layers) + KD} \\
         \midrule
         S2.1: Prune layers at random & 79.4 & 77.1 & 87.6 & 87.6 & 88.7 & 90.3 \\
         S2.2: Prune last N layers & 80.6 & 78.8 & \bf 92.1 & \bf 91.2 & \bf 88.8 & \bf 91.2 \\
         S2.3: Prune first N layers & 77.8 & 76.4 & 90.8 & 89.2 & 78.9 & 83.6 \\
         S2.4: Prune every second layer & \bf 81.0 & \bf 79.6 & 89.9 & 88.5 & 88.6 & 89.4 \\
         S2.5: Prune layers with mimimum pair-wise distance & 75.6 & 72.6 & 91.5 & 90.5 & 85.6 & 87.3 \\
         \midrule
         \multicolumn{7}{c}{Input Model: S2.2 --$>$  S3: Pruning Width (Heads + FF) + KD} \\
         \midrule
         S3.1: Prune heads and FF at random & 77.9 & 75.8 & 90.6 & 89.3 & 82.6 & 87.1 \\ 
         S3.2: Prune heads and FF based on the exp. sensitivity(L)  & \bf 78.0 & \bf 76.1 & \bf 91.6 & \bf 90.7 & \bf 89.5 & \bf 92.3 \\ 
    \end{tabular}
    }
    \caption{Preliminiary Experiments to determine which settings produce the most consistent results.}
    \label{tab:preliminiary_experiments}
    \vspace{-3mm}
\end{table*}

\section{Experimentail Details}
\label{sec:appendix_a}

\citet{devlin-etal-2019-bert} suggested a hyperparameter tuning, that was adopted in many papers \cite{Lee_2019,beltagy-etal-2019-scibert,alsentzer-etal-2019-publicly,sung-etal-2019-pre}. This hyper-parameter tuning included a light grid search in learning rate $\in$ \{2e-5, 3e-5, 4e-5, 5e-5\}, the number of training epochs $\in$  \{3, 4\}, and the batch size $\in$ \{16, 32\} with a fixed dropout rate of 0.1. In our research, we tune each variation of our model based on a grid-search of learning rate on the following range $\in$  \{1e-4, 3e-4, 1e-5, 3e-5, 5e-5, 1e-6\}. The batch size is fixed to 16, and the dropout rate at 0.1. The max sequence length is fixed to 512 for MultiEURLEX and ContractNLI, 256 for ContractNER and 128 for Contract-Obligations and UNFAIR-ToS based on the training subset statistics. Lastly, \citet{chalkidis-etal-2020-legalbert} found that some models may underfit for 4 epochs. Hence, following their work, we use early stopping based on validation loss up to 20 maximum train epochs with a patience of 3 epochs.

In every incremental step of GC where knowledge distillation is applied, learning rate, temperature and $a$ (the original and distillation loss weighing) are tuned using grid search. in the hyper-parameter spaces of [1e-5, 3e-5, 5-e5, 7e-5, 1e-4], [1, 5, 10, 15] and [0.1, 0.3, 0.6], respectively.

\section{Additional Results}
\label{sec:appendix_b}

\subsection{Additional Pre-training Results}
\label{app:additional_pre-training_results}

This section provides additional results regarding the pre-training process. Table~\ref{tab:corpora_tokens} displays the number of tokens that were included in contracts and regulations for each language. It should be noted that during the pre-training, 100\% of regulations and 20\% of translated contracts (100\% of English contracts used) were used. Finally, Table~\ref{tab:detailed_mlm} presents the evaluation loss and accuracy scores of the pre-training of the masked language models. This table provides the overall scores, along with scores for each document type, language and document type/language.

\begin{table*}[t]
    \centering
    \resizebox{0.8\textwidth}{!}{
    \begin{tabular}{l|c|c|c} 
        \multirow{2}{*}{\bf Task} & \multicolumn{1}{c|}{\bf Reduction of} & \multicolumn{1}{c|}{\bf Removed Params} & \multicolumn{1}{c}{\bf Removed Params} \\
        & {\bf Tokens} & {\bf (C-XLM Large)} & {\bf (C-XLM Base)} \\
         \midrule
         MultiEURLEX & 2.52\% & 1,653,760 &826,880 \\
         Obligations & 26.98\% & 17,679,360 & 8,839,680 \\
         UNFAIR-ToS & 22.54\% & 14,774,272 & 7,387,136 \\
         ContractNLI & 31.83\% & 20,858,880 & 10,429,440 \\
         ContractNER & 24.17\% & 15,839,232 & 7,919,616 \\
    \end{tabular}
    }
    \medskip
    \caption{Percentage of usable tokens and parameter reduction after vocabulary pruning for each task.}
    \label{tab:vocab_reduction}
\end{table*}

\begin{table*}[t]
\centering
\resizebox{\textwidth}{!}{
\begin{tabular}{l|c|c|c|c|c}
\bf Model  & \bf MultiEURLEX & \bf UNFAIR-ToS & \bf CNLI & \bf Obligations & \bf ContractNER \\
\midrule
XLM-R (Base)  &  1e-5           &      3e-5      &  1e-5    &      1e-5       &      5e-5       \\
XLM-R (Large) &      3e-5       &      1e-5      &   1e-5   &     1e-6        &     5e-5        \\
\midrule
C-XLM (Tiny)  & 1e-4        & 3e-4       & 1e-4 & 1e-4        & 1e-4        \\
C-XLM (Small) & 1e-4        & 5e-5       & 1e-4 & 5e-5        & 5e-5        \\
C-XLM (Base)  & 5e-5        & 3e-5       & 5e-5 & 3e-5        & 3e-5        \\
C-XLM (Large) & 5e-5        & 5e-5       & 3e-5 & 1e-5        & 5e-5       \\
\bottomrule
\end{tabular}
}
\caption{Optimal Learning Rates per downstream task across all models.}
\label{tab:learning_rates}
    \vspace{-3mm}
\end{table*}

\subsection{Model Compression Preliminary Experiments}
\label{app:preliminary_experiments}
Before each model was compressed, some preliminary experiments were conducted to determine which setting in each compression step produces the most consistent results. 

\begin{figure}[t]
    \centering
    \includegraphics[width=\columnwidth]{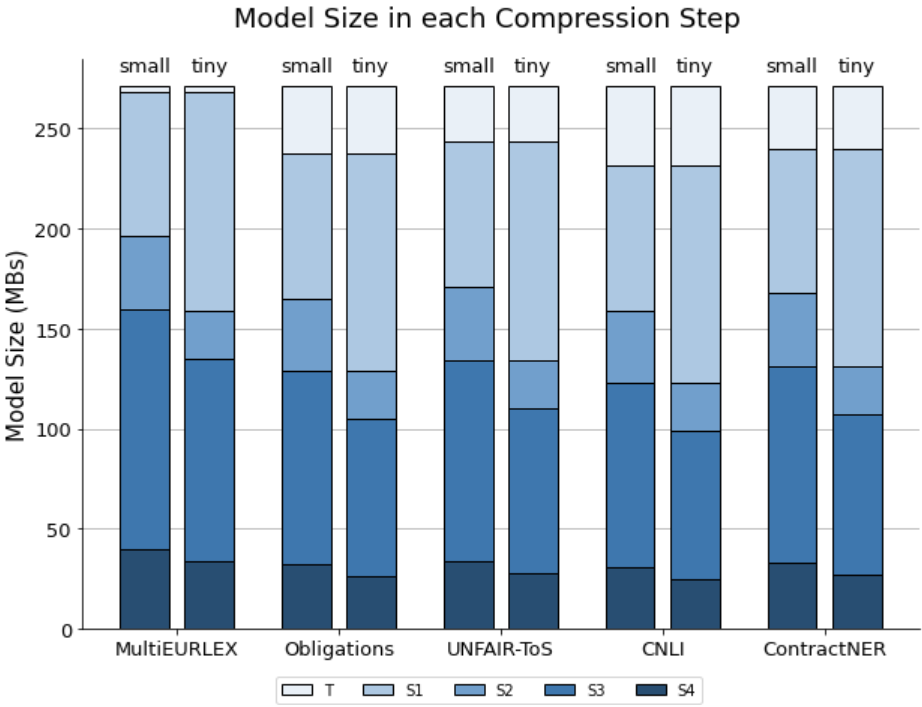}
        \vspace{-5mm}
    \caption{Model size (MB) in each compression step (S) in relevance to the original model, C-XLM (large).}
    \label{fig:compression_sizesb}
    \vspace{-4mm}
\end{figure}

For the first step (Vocabulary Reduction), our proposed method, where the model's vocabulary is pruned based on information from the training data and the tokenizer's merges, was compared with a random baseline, i.e., the vocabulary is randomly pruned. In both settings, the exact same percentage of tokens were kept, and as expected, the random baseline led to performance deterioration.

For the second step (Depth Pruning), 9 out of the 12 transformer blocks of the base (language) model were removed and the pruned model was trained using knowledge distillation. Five different settings were tested. Encoder blocks were pruned by: (i) pruning blocks at random, (ii) pruning the last 9 blocks, (iii) pruning the first 9 blocks, (iv) pruning every second block, and (v) pruning blocks with the minimum pair-wise distance. For setting (v), the mean absolute error and the cosine similarity of the CLS tokens of each encoder block were used as the metrics (except for the ContractNER task, were the average of all tokens was used instead of the CLS one). We found that among all settings, copying the weights of the first encoder blocks of the original pre-trained language model produced the most consistent results.

For the third step (Width Pruning), the model of the second setting from the second step was pruned down to 12 attention heads (in total) and 512 neurons in the intermediate fully-connected layers. Two settings were examined: random pruning and pruning based on the expected loss when each attention head/neuron was iteratively removed \cite{michel2019sixteen,prasanna2020bert}. Just like in step 1, the random baseline performed worse. Results are summarized in Table ~\ref{tab:preliminiary_experiments}.

\subsection{Additional Fine-tuning Results}
\label{app:additional_fine-tuning_results}

This section presents some additional results regarding the fine-tuning process. Table~\ref{tab:learning_rates} displays the optimal learning rates for each model variation and baseline that were used during the fine-tuning process for every different task. Each one was selected through grid search, between learning rates $\in$  \{1e-4, 3e-4, 1e-5, 3e-5, 5e-5, 1e-6\}. We observe that smaller models favor larger learning rates, i.e., 1e-4 and 5e-5 in most cases, while larger models favor smaller learning rates, i.e., 1e-5 and 3e-5. Additionally, the upper parts of Tables~\ref{tab:eurl_per_lang_micro} and \ref{tab:eurl_per_lang_macro} present the \microf and \macrof scores of the fine tuned models per language, respectively, for MultiEURLEX task. Lastly, the upper part of Table~\ref{tab:unfair_per_lang} presents the Mean Absolute Error (MAE) and accuracy scores of the fine-tuned models, for UNFAIR-ToS task per language.

\subsection{Additional Compression Results}
\label{app:additional_experiments}

In this section, some additional experimental results are presented. First, the percentage of usable tokens and the parameter reduction of both large and base models from the vocabulary pruning step for each task can be found in Table ~\ref{tab:vocab_reduction}. In Table~\ref{tab:compression_size_speed}, the model size (in MBs) and the average inference time (in seconds) of a 32-batch (on CPU) across all incremental compression steps and baselines are presented. Inference benchmarking was conducted using a modern mid-range \textsc{Ryzen 7 4700u}. In Table~\ref{tab:base_teacher}, model compression results can be found, across all down-stream task, when the Base C-XLM is used as a teacher. Lastly, Tables~\ref{tab:eurl_per_lang_micro}, ~\ref{tab:eurl_per_lang_macro} and ~\ref{tab:unfair_per_lang} summarize the \microf, \macrof of MultiEURLEX and results of UNFAIR-ToS tasks, across all languages for each incremental compression step and baselines.

\begin{table*}[t]
    \centering
    \resizebox{\textwidth}{!}{
    \begin{tabular}{l|cc|cc|cc|cc|cc}
         \multirow{2}{*}{\bf Model} & \multicolumn{2}{c|}{\bf MultiEURLEX} & \multicolumn{2}{c|}{\bf UNFAIR-ToS} & \multicolumn{2}{c|}{\bf CNLI} & \multicolumn{2}{c|}{\bf Obligations} & \multicolumn{2}{c}{\bf ContractNER}  \\
         & \microf & \macrof & Acc. & MAE & \microf & \macrof & \microf & \macrof & \microf & \macrof \\
         \midrule
         \multicolumn{11}{c}{Top Bound - Performance ``Ceiling''} \\
         \midrule
         C-XLM (large) & 78.4 & 65.4 & 89.7 & 0.14 & 85.3 & 83.0 & 91.8 & 90.6 & 93.2 & 94.6 \\
         \midrule
         Step 0 (TA-KD) & 75.2 & 63.0 & 88.6 & 0.16 & 84.6 & 82.2 & 92.8 & 91.8 & 93.7 & 94.9\\
         \midrule
         \multicolumn{11}{c}{Gradual Compression --- Reference C-XLM (small)} \\
         \midrule
         Step 1 (VP+KD) & 75.1 & 62.9 & 88.5 & 0.18 & 84.9 & 82.9 & 92.8 & 91.8 & 93.6 & 94.9 \\
         Step 2 (DP+KD) & 74.4 & 54.3 & 83.5 & 0.22 & 84.0 & 81.8 & 92.6 & 91.6 & 92.4 & 93.2 \\
         Step 3 (WP+KD) & 73.8 & 60.8 & 83.5 & 0.21 & 84.5 & 83.0 & 92.3 & 91.4 & 92.1 & 93.2 \\
         Step 4 (GO+Q) & \bf 74.2 & \bf 60.4 & 83.7 & 0.21 & \bf 84.5 & \bf 83.1 & \bf 92.2 & \bf 91.3 & \bf 92.2 & \bf 93.3 \\
         \midrule
         C-XLM (small) (FT) & 72.3 & 54.7 & \bf 85.4 & \bf 0.20 & 79.7 & 77.0 & 90.4 & 89.0 & 90.1 & 92.4  \\
         C-XLM (small) (KD) & 73.3 & 54.7 & 81.1 & 0.25 & 80.2 & 78.1 & 90.1 & 89.1 & 91.0 & 93.1\\
         \midrule
         \multicolumn{11}{c}{Gradual Compression --- Reference C-XLM (tiny)} \\
         \midrule
         Step 1 (VP+KD) & 75.1 & 62.9 & 88.5 & 0.18 & 84.9 & 82.9 & 92.8 & 91.8 & 93.6 & 94.9 \\
         Step 2 (DP+KD) & 72.6 & 58.2 & 80.6 & 0.24 & 80.9 & 79.3 & 91.1 & 90.0 & 89.7 & 92.1 \\
         Step 3 (WP+KD) & 72.5 & 58.0 & 79.9 & 0.25 & 80.9 & 79.4 & 91.6 & 90.5 & 87.6 & 90.4 \\
         Step 4 (GO+Q) & \bf 73.2 & \bf 57.0 & \bf 79.6 & \bf 0.25 & \bf 80.7 & \bf 79.2 & \bf 91.9 & \bf 90.7 & 87.6 & \bf 90.2 \\
         \midrule
         C-XLM (tiny) (FT) & 66.5 & 46.1 & 78.2 & 0.27 & 70.2 & 69.2 & 88.7 & 87.4 & 87.2 & 89.3 \\
         C-XLM (tiny) (VP+KD) & 64.0 & 42.0 & 76.7 & 0.30 & 75.3 & 74.3 & 89.1 & 88.1 & \bf 87.7 & 90.1 \\
         \bottomrule
    \end{tabular}
    }
    \caption{Model compression results across down-stream tasks. We report the model's performance across the incremental compression steps (S) presented in Section~\ref{sec:compression_methodology}. We also report the performance for two baselines: (a) fine-tuning the reference pre-trained C-XLM model (FT), and (b) Knowledge Distillation and Vocabulary Pruning. where the student is the reference pre-trained C-XLM (KD).}
    \label{tab:compression_perfomance_steps}
\end{table*}

\begin{table*}[t]
    \centering
    \resizebox{\textwidth}{!}{
    \begin{tabular}{l|cc|cc|cc|cc|cc}
         \multirow{2}{*}{\bf Model} & \multicolumn{2}{c|}{\bf MultiEURLEX} & \multicolumn{2}{c|}{\bf UNFAIR-ToS} & \multicolumn{2}{c|}{\bf CNLI} & \multicolumn{2}{c|}{\bf Obligations} & \multicolumn{2}{c}{\bf ContractNER}  \\
         & \microf & \macrof & Acc. & MAE & \microf & \macrof & \microf & \macrof & \microf & \macrof \\
         \midrule
         \multicolumn{11}{c}{Top Bound - Performance ``Ceiling''} \\
         \midrule
         C-XLM (Base) & 75.3 & 59.4 & 87.3 & 0.18 & 84.0 & 82.1 & 91.2 & 90.4 & 92.9 & 93.8\\
         \midrule
         \multicolumn{11}{c}{Gradual Compression --- Reference C-XLM (small)} \\
         \midrule
         Step 1 (VP+KD) & 74.9 & 60.3 & 88.2 & 0.17 & 84.8 & 82.9 & 91.7 & 90.6 & 93.2 & 94.1\\
         Step 2 (DP+KD) & 74.4 & 59.3 & 82.7 & 0.23 & 84.2 & 82.4 & 91.3 & 90.2 & 91.6 & 91.7\\
         Step 3 (WP+KD) & 73.8 & 61.5 & 83.2 & 0.23 & 84.8 & 83.7 & 92.8 & 91.5 & 92.7 & 94.1\\
         Step 4 (GO+Q) & \bf 73.7 & \bf 61.7 & 83.2 & 0.23 & \bf 84.5 &  \bf 83.1 & \bf 92.7 & \bf 91.4 & \bf 92.7 & \bf 93.7\\
         \midrule
         C-XLM (small) (FT) & 69.9 & 51.7 & \bf 85.4 & \bf 0.20 & 79.7 & 77.0 & 90.4 & 89.0 & 90.1 & 92.4\\
         C-XLM (small) (KD) & 72.3 & 54.7 & 82.7 & 0.23 & 80.0 & 78.4 & 90.3 & 89 & 92.0 & 92.4\\
         \midrule
         \multicolumn{11}{c}{Gradual Compression --- Reference C-XLM (tiny)} \\
         \midrule
         Step 1 (VP+KD) &  74.9 & 60.3 & 88.2 & 0.17 & 84.8 & 82.9 & 91.7 & 90.6 & 93.2 & 94.1\\
         Step 2 (DP+KD) & 71.2 & 55.7 & 81.8 & 0.22 & 79.1 & 77.0 & 92.1 & 91.2 & 89.0 & 90.6\\
         Step 3 (WP+KD) & 70.6 & 55.8 & 80.8 & 0.24 & 78.0 & 76.1 & 91.6 & 90.7 & 89.8 & 92.5\\
         Step 4 (GO+Q) & \bf 70.4 & \bf 54.4 & \bf 80.8 & \bf 0.24 & \bf 78.5 & \bf 76.8 & \bf 91.6 & \bf 90.7 & \bf 89.5 & \bf 92.3\\
         \midrule
         C-XLM (tiny) (FT) &  66.5 & 46.1 & 78.2 & 0.27 & 70.2 & 69.2 & 88.7 & 87.4 & 87.2 & 89.3\\
         C-XLM (tiny) (KD) & 66.3 & 45.3 & 75.3 & 0.31 & 74.6 & 72.2 & 86.8 & 85.6 & 87.7 & 89.9\\
         \bottomrule
    \end{tabular}
    }
    \caption{Model compression results across down-stream tasks when Base C-XLM is used as a teacher. We report the model's performance across the incremental compression steps (S) presented in Section~\ref{sec:compression_methodology}. We also report the performance for two baselines: (a) fine-tuning the reference pre-trained C-XLM model (FT), and (b) Knowledge Distillation and Vocabulary Pruning. where the student is the reference pre-trained C-XLM (KD).}
    \label{tab:base_teacher}
\end{table*}

\begin{table*}[t]
    \centering
    \resizebox{\textwidth}{!}{
    \begin{tabular}{l|cc|cc|cc|cc|cc}
         \multirow{2}{*}{\bf Model} & \multicolumn{2}{c|}{\bf MultiEURLEX} & \multicolumn{2}{c|}{\bf UNFAIR-ToS} & \multicolumn{2}{c|}{\bf CNLI} & \multicolumn{2}{c|}{\bf Obligations} & \multicolumn{2}{c}{\bf ContractNER}  \\
         & Size &  Time & Size & Time & Size & Time & Size &  Time & Size &  Time \\
         \midrule
         \multicolumn{11}{c}{Top Bound - Performance ``Ceiling''} \\
         \midrule
         C-XLM (large) & 1,409 & 99.3 & 1,409 & 21.5 & 1,409 & 116.0 & 1,409 & 21.3 & 1409 & 45.6\\
         \midrule
         \multicolumn{11}{c}{Gradual Compression --- Reference C-XLM (tiny)} \\
         \midrule
         Step 1 (VP+KD) & 268 & 16.1 & 243 & 3.6 & 232 & 20.2 & 238 & 3.3 & 240 & 7.4\\
         Step 2 (DP+KD) & 159 & 4.1 & 134 & 0.8 & 123 & 4.1 & 129 & 0.8 & 131 & 1.8\\
         Step 3 (WP+KD) & 135 & 1.8 & 110 & 0.4 & 99 & 1.8 & 105 & 0.3 & 107 & 0.8\\
         Step 4 (GO+Q) & \bf 34 & \bf 0.8 & \bf 28 & \bf 0.1 & \bf 25 & \bf 0.8 & \bf 26 & \bf 0.1 & \bf 27 & \bf 0.3\\
         \midrule
         C-XLM (tiny) (FT) & 35 & 1.3 & 35 & 0.2 & 35 & 2.0 & 35 & 0.2 & 35 & 0.5\\
         C-XLM (tiny) (KD) & \bf 34 & 1.3 & \bf 28 & 0.2 & \bf 25 & 1.3 & \bf 26 & 0.2 & \bf 27 & 0.5\\
         \midrule
         \multicolumn{11}{c}{Gradual Compression --- Reference C-XLM (small)} \\
         \midrule
         Step 1 (VP+KD) & 268 & 16.1 & 243 & 3.6 & 232 & 20.2 & 238 & 3.3 & 240 & 7.4\\
         Step 2 (DP+KD) & 196 & 8.3 & 171 & 1.8 & 159 & 8.3 & 165 & 1.6 & 168 & 3.5\\
         Step 3 (WP+KD) & 160 & 4.4 & 110 & 0.9 & 123 & 4.4 & 129 & 0.9 & 131 & 1.9\\
         Step 4 (GO+Q) & \bf 40 & \bf 1.8 & \bf 34 & \bf 0.3 & \bf 31 & \bf 1.8 & \bf 32 & \bf 0.3 & \bf 33 & \bf 0.7\\
         \midrule
         C-XLM (small) (FT) & 81 & 3.2 & 81 & 0.6 & 81 & 4.1 & 81 & 0.5 & 81 & 1.4\\
         C-XLM (small) (KD) & 80 & 3.2 & 67 & 0.6 & 61 & 3.2 & 65 & 0.5 & 66 & 1.3\\
    \end{tabular}
    }
    \caption{Model compression results across down-stream tasks. We report the model's size in MBs and average inference time, in seconds, of a 32-batch across the incremental compression steps (S) presented in Section~\ref{sec:compression_methodology}. We also report the performance for two baselines: (a) fine-tuning the reference pre-trained C-XLM model (FT), and (b) Knowledge Distillation and Vocabulary Pruning. where the student is the reference pre-trained C-XLM (KD).}
    \label{tab:compression_size_speed}
\end{table*}

\begin{table*}[t]
    \centering
    \resizebox{\textwidth}{!}{
    \begin{tabular}{l|ccccccccc|c}
         \multirow{2}{*}{\bf Model} & \multicolumn{10}{c}{\bf MultiEURLEX per Language \microf} \\
         & EN & FR & DE & NL & IT & ES & PT & PL & EL & \microf \\
         \midrule
         \multicolumn{11}{c}{Fine-tuned Models} \\
         \midrule
         XLM-R (large) & 80.6 & 78.8 & 76.1 & 76.1 & 78.5 & 80.1 & 75.7 & 75.8 & 78.7 & 77.8\\
         XLM-R (base) & 77.6 & 76.9	& 73.7	& 74.3	& 76.3	& 75.0	& 75.7	& 73.2	& 75.6	& 75.3\\
         \midrule
         C-XLM (large) & 80.5 & 77.7 & 76.4 & 76.8 & 77.4 & 79.7 & 78.3 & 76.9 & 82.0 & 78.4 \\
         C-XLM (base) & 77.5 & 76.4	& 72.3	& 74.1	& 75.6	& 77.4	& 75.0	& 72.9	& 76.5	& 75.3\\
         C-XLM (small) & 69.0 & 65.7 & 65.1 & 62.2 & 66.4 & 66.4 & 66.6 & 65.3 & 71.6 & 72.3\\
         C-XLM (tiny)  & 69.0 & 65.7 & 65.1 & 62.2 & 66.4 & 66.4 & 66.6 & 65.3 & 71.6 & 66.5 \\
         \midrule
         \multicolumn{11}{c}{Gradual Compression --- Reference C-XLM (tiny)} \\
         \midrule
         Step 1 (VP+KD) & 78.0 & 75.9 & 74.0 & 73.1 & 75.2 & 78.5 & 75.4 & 73.4 & 72.0 & 75.1 \\
         Step 2 (DP+KD) & 75.3 & 72.3 & 67.2 & 71.4 & 73.9 & 75.9 & 73.3 & 72.3 & 71.9 & 72.6 \\
         Step 3 (WP+KD) &  74.1 & 70.7 & 69.2 & 71.9 & 74.4 & 74.2 & 72.9 & 71.8 & 73.1 & 72.5\\
         Step 4 (GO+Q) &  74.2 & 70.8 & 70.9 & 72.8 & 74.7 & 74.2 & 73.7 & 72.5 & 73.5 & 73.2\\
         \midrule
         C-XLM (tiny) (KD) &  64.9 & 63.3 & 60.8 & 63.2 & 66.3 & 63.1 & 63.8 & 65.0 & 65.7 & 64.0\\
         \midrule
         \multicolumn{11}{c}{Gradual Compression --- Reference C-XLM (small)} \\
         \midrule
         Step 1 (VP+KD) &  78.0 & 75.9 & 74.0 & 73.1 & 75.2 & 78.5 & 75.4 & 73.4 & 2.0 & 75.1\\
         Step 2 (DP+KD) &  76.5 & 75.6 & 71.0 & 71.4 & 75.9 & 77.4 & 75.5 & 73.0 & 73.1 & 74.4\\
         Step 3 (WP+KD) & 75.8 & 75.2 & 70.4 & 70.5 & 74.4 & 75.6 & 73.9 & 74.2 & 74.7 & 73.8\\
         Step 4 (GO+Q) & 75.9 & 75.7 & 71.7 & 71.4 & 74.8 & 76.1 & 74.1 & 73.7 & 75.5 & 74.2\\
         \midrule
         C-XLM (small) (KD) & 74.2 & 73.2 & 71.7 & 72.7 & 76.4 & 73.3 & 75.5 & 71.6 & 70.8 & 73.3\\
    \end{tabular}
    }
    \caption{Model compression results for the MultiEURLEX task. We report the model's per-language \microf across the incremental compression steps (S) presented in Section~\ref{sec:compression_methodology}. We also report the performance for two baselines: (a) fine-tuning the reference pre-trained C-XLM model (FT), and (b) Knowledge Distillation and Vocabulary Pruning. where the student is the reference pre-trained C-XLM (KD).}
    \label{tab:eurl_per_lang_micro}
\end{table*}

\begin{table*}[t]
    \centering
    \resizebox{\textwidth}{!}{
    \begin{tabular}{l|ccccccccc|c}
         \multirow{2}{*}{\bf Model} & \multicolumn{10}{c}{\bf MultiEURLEX per Language \macrof} \\
         & EN & FR & DE & NL & IT & ES & PT & PL & EL & \macrof \\
         \midrule
         \multicolumn{11}{c}{Fine-tuned Models} \\
         \midrule
         XLM-R (large) & 64.3 & 67.3 & 60.3 & 61.9 & 58.1 & 63.6 & 57.1 & 60.0 & 63.5 & 63.8\\
         XLM-R (base) & 56.2 & 54.7	& 50.9	& 53.6	& 49.5	& 52.4	& 52.2	& 49.8	& 52.0	& 53.2\\
         \midrule
         C-XLM (large) & 66.8 & 63.2 & 63.0 & 63.9 & 55.0 & 67.1 & 60.6 & 61.6 & 70.7 & 65.4 \\
         C-XLM (base) & 59.6 & 58.8	& 55.1	& 59.7	& 54.0	& 59.1	& 59.8	& 57.1	& 60.3	& 59.4\\
         C-XLM (small) & 55.9 & 52.7 & 50.8 & 55.8 & 52.6 & 57.0 & 55.6 & 50.8 & 56.2 & 54.7\\
         C-XLM (tiny)  & 50.1 & 43.7 & 47.7 & 42.6 & 38.3 & 46.8 & 47.6 & 41.9 & 44.1 & 46.1\\
         \midrule
         \multicolumn{11}{c}{Gradual Compression --- Reference C-XLM (tiny)} \\
         \midrule
         Step 1 (VP+KD) & 64.3 & 64.3 & 60.8 & 60.8 & 53.2 & 66.6 & 63.3 & 59.0 & 56.0 & 62.9\\
         Step 2 (DP+KD) & 58.4 & 56.9 & 51.0 & 55.6 & 56.4 & 61.0 & 55.5 & 56.2 & 54.9 & 58.2\\
         Step 3 (WP+KD) & 56.1 & 62.2 & 55.5 & 53.7 & 53.6 & 59.7 & 55.5 & 54.3 & 53.9 & 58.0\\
         Step 4 (GO+Q) & 54.8 & 56.0 & 54.6 & 54.2 & 53.5 & 58.8 & 54.0 & 54.4 & 51.1 & 57.0\\
         \midrule
         C-XLM (tiny) (KD) & 42.2 & 41.1 & 40.9 & 38.7 & 41.5 & 40.9 & 41.5 & 41.2 & 39.4 & 42.0\\
         \midrule
         \multicolumn{11}{c}{Gradual Compression --- Reference C-XLM (small)} \\
         \midrule
         Step 1 (VP+KD) & 64.3 & 64.3 & 60.8 & 60.8 & 53.2 & 66.6 & 63.3 & 59.0 & 56.0 & 62.9\\
         Step 2 (DP+KD) & 58.2 & 55.8 & 49.9 & 56.0 & 50.2 & 56.5 & 53.6 & 49.8 & 50.8 & 54.3\\
         Step 3 (WP+KD) & 61.9 & 60.9 & 56.2 & 59.1 & 53.8 & 61.8 & 55.5 & 61.1 & 58.4 & 60.8\\
         Step 4 (GO+Q) & 61.5 & 60.8 & 57.3 & 59.8 & 54.7 & 61.6 & 55.4 & 57.6 & 58.8 & 60.4\\
         \midrule
         C-XLM (small) (KD) & 56.9 & 54.3 & 52.8 & 56.3 & 54.3 & 55.1 & 55.8 & 48.6 & 47.8 & 54.7\\
    \end{tabular}
    }
    \caption{Model compression results for the MultiEURLEX task. We report the model's per-language \macrof across the incremental compression steps (S) presented in Section~\ref{sec:compression_methodology}. We also report the performance for two baselines: (a) fine-tuning the reference pre-trained C-XLM model (FT), and (b) Knowledge Distillation and Vocabulary Pruning. where the student is the reference pre-trained C-XLM (KD).}
    \label{tab:eurl_per_lang_macro}
\end{table*}

\begin{table*}[t]
    \centering
    \resizebox{\textwidth}{!}{
    \begin{tabular}{l|cc|cc|cc|cc|cc}
         \multirow{3}{*}{\bf Model} & \multicolumn{10}{c}{\bf UNFAIR-ToS per Language \macrof} \\ 
         & \multicolumn{2}{c|}{EN} & \multicolumn{2}{c|}{PL} & \multicolumn{2}{c|}{IT} & \multicolumn{2}{c|}{DE} & \multicolumn{2}{c}{Total}\\
         & MAE & \microf & MAE & \microf & MAE & \microf & MAE & \microf & MAE & \microf \\
         \midrule
         \multicolumn{11}{c}{Fine-tuned Models} \\
         \midrule        
         XLM-R (large) & 0.16 & 89.3 & 0.19 & 87.2 & 0.14 & 91.2 & 0.15 & 88.3 & 0.16 & 89.0 \\
         XLM-R (base) & 0.14 & 90.3 & 0.22 & 81.7 & 0.15 & 88.2 & 0.17 & 86.4 & 0.17 & 86.6\\
         \midrule
         C-XLM (large) & 0.13 & 90.3 & 0.17 & 88.1 & 0.11 & 92.2 & 0.15 & 88.3 & 0.14 & 89.7 \\
         C-XLM (base) & 0.17 & 87.4 & 0.22 & 83.5 & 0.13 & 91.2 & 0.19 & 87.4 & 0.18 & 87.3\\
         C-XLM (small) & 0.17 & 85.4 & 0.24 & 80.7 & 0.16 & 90.2 & 0.21 & 85.4 & 85.4 & 0.20\\ 
         C-XLM (tiny)  & 0.27 & 82.5 & 0.28 & 76.1 & 0.24 & 79.4 & 0.30 & 74.8 & 78.2 & 0.27\\
         \midrule
         \multicolumn{11}{c}{Gradual Compression --- Reference C-XLM (tiny)} \\
         \midrule
         Step 1 (VP+KD) & 0.13 & 92.2 & 0.21 & 86.2 & 0.14 & 90.2 & 0.22 & 85.4 & 0.18 & 88.5\\
         Step 2 (DP+KD) & 0.23 & 80.6 & 0.24 & 78.0 & 0.22 & 84.3 & 0.27 & 79.6 & 0.24 & 80.6\\
         Step 3 (WP+KD) & 0.24 & 80.6 & 0.24 & 80.7 & 0.23 & 82.4 & 0.28 & 75.7 & 0.25 & 79.9\\
         Step 4 (GO+Q) & 0.24 & 80.6 & 0.24 & 80.7 & 0.24 & 82.4 & 0.28 & 75.7 & 0.25 & 79.6\\
         \midrule
         C-XLM (tiny) (KD) & 0.32 & 76.7 & 0.30 & 75.2 & 0.30 & 76.5 & 0.27 & 78.6 & 0.30 & 76.7\\
         \midrule
         \multicolumn{11}{c}{Gradual Compression --- Reference C-XLM (small)} \\
         \midrule
         Step 1 (VP+KD) & 0.13 & 92.2 & 0.21 & 86.2 & 0.14 & 90.2 & 0.22 & 85.4 & 0.18 & 88.5\\
         Step 2 (DP+KD) & 0.20 & 83.5 & 0.25 & 80.7 & 0.21 & 84.3 & 0.21 & 85.4 & 0.22 & 83.5\\
         Step 3 (WP+KD) & 0.20 & 84.5 & 0.21 & 84.4 & 0.22 & 81.4 & 0.21 & 83.5 & 0.21 & 83.5\\
         Step 4 (GO+Q) & 0.20 & 84.5 & 0.20 & 84.4 & 0.22 & 82.4 & 0.21 & 83.5 & 0.21 & 83.7\\
         \midrule
         C-XLM (small) (KD) & 0.25 & 81.6 & 0.28 & 78.0 & 0.22 & 83.3 & 0.26 & 81.6 & 0.25 & 81.1\\
    \end{tabular}
    }
    \caption{Model compression \macrof for the UNFAIR-ToS task. We report the per-language model's \microf and MAE across the incremental compression steps (S) presented in Section~\ref{sec:compression_methodology}. We also report the performance for two baselines: (a) fine-tuning the reference pre-trained C-XLM model (FT), and (b) Knowledge Distillation and Vocabulary Pruning. where the student is the reference pre-trained C-XLM (KD).}
    \label{tab:unfair_per_lang}
\end{table*}

\begin{table*}[t]
\centering
\resizebox{\textwidth}{!}{
\begin{tabular}{l | ll | ll |ll |ll |ll |ll}
\bf Model &  \multicolumn{8}{c|}{\bf C-XLM} & \multicolumn{4}{c}{\bf XLM-R} \\
\midrule
\multirow{2}{*}{\bf Corpus Subset}& \multicolumn{2}{c |}{\bf Tiny}  & \multicolumn{2}{c |}{\bf Small}  & \multicolumn{2}{c |}{\bf Base}   & \multicolumn{2}{c |}{\bf Large}  & \multicolumn{2}{c |}{\bf Base} & \multicolumn{2}{c }{\bf Large}\\ 
& \bf Loss & \bf Acc. & \bf Loss & \bf Acc. & \bf Loss & \bf Acc. & \bf Loss & \bf Acc. & \bf Loss & \bf Acc. & \bf Loss & \bf Acc.\\
\midrule
Regulations (EN) & 2.46      & 54.2\%     & 1.62     & 67.1\%      & 1.11     & 76.3\%     & 0.97      & 80.0\%     & 1.36           & 71.2\%           & 1.07           & 76.3\%            \\
Regulations (EL) & 1.89      & 61.0\%     & 1.24     & 72.9\%      & 0.85     & 80.5\%     & 0.74      & 83.9\%     & 0.93           & 79.2\%           & 0.68           & 84.3\%            \\
Regulations (DE) & 2.52     & 52.3\%     & 1.60     & 67.3\%      & 1.02     & 77.6\%     & 0.84      & 82.1\%     & 1.20            & 74.2\%           & 0.89           & 80.0\%            \\
Regulations (FR) & 1.84     & 62.3\%     & 1.16      & 74.8\%      & 0.75     & 82.8\%     & 0.65      & 86.0\%     & 1.01           & 77.8\%           & 0.77           & 82.5\%            \\
Regulations (ES) & 2.01     & 59.7\%     & 1.31     & 72.1\%      & 0.88     & 80.2\%     & 0.78      & 83.3\%     & 1.19           & 74.4\%           & 0.93           & 79.0\%            \\
Regulations (NL) & 2.43     & 54.0\%     & 1.54     & 68.4\%      & 1.00     & 78.1\%     & 0.85      & 82.1\%     & 1.25           & 73.8\%           & 0.91           & 79.8\%            \\
Regulations (IT) & 2.06     & 58.5\%     & 1.32     & 71.8\%      & 0.88     & 80.2\%     & 0.77      & 83.5\%     & 1.16           & 75.0\%           & 0.88           & 80.2\%            \\
Regulations (PL) & 2.23      & 57.2\%     & 1.44     & 70.3\%      & 0.94     & 79.2\%     & 0.75      & 83.3\%     & 1.08           & 76.8\%           & 0.80           & 82.1\%            \\
Regulations (PT) & 2.20     & 57.2\%     & 1.43     & 70.7\%      & 0.98     & 79.0\%     & 0.87      & 82.3\%     & 1.23           & 73.5\%           & 0.94           & 78.8\%            \\
\midrule
Regulations (All) & 2.36     & 55.1\%     & 1.53     & 68.9\%      & 1.03     & 77.7\%     & 0.90      & 81.4\%     & 1.18           & 74.6\%           & 0.90           & 79.8\%            \\
\midrule
Contracts (EN)   & 1.96     & 58.0\%     & 1.15     & 73.7\%      & 0.66     & 84.2\%     & 0.45      & 89.1\%     & 1.24           & 73.0\%           & 0.93           & 78.8\%            \\
Contracts (EL)   & 2.11     & 57.2\%     & 1.41     & 70.0\%      & 0.99     & 78.0\%     & 0.84      & 81.8\%     & 1.07            & 76.8\%           & 0.87           & 80.9\%            \\
Contracts (DE)   & 2.62     & 50.0\%     & 1.65     & 66.2\%      & 1.07     & 76.5\%     & 0.89      & 80.7\%     & 1.27           & 73.0\%           & 1.04           & 77.4\%            \\
Contracts (FR)   & 1.99     & 59.9\%     & 1.23     & 73.9\%      & 0.78     & 82.5\%     & 0.65      & 85.9\%     & 1.09           & 76.8\%           & 0.88           & 80.7\%            \\
Contracts (ES)   & 2.28      & 54.7\%     & 1.45      & 69.2\%      & 0.95     & 78.6\%     & 0.80      & 82.5\%     & 1.32           & 72.4\%           & 1.08           & 76.4\%            \\
Contracts (NL)   & 2.64     & 49.7\%     & 1.69      & 65.2\%      & 1.12     & 75.5\%     & 0.93     & 79.9\%     & 1.43           & 70.6\%           & 1.15           & 75.5\%            \\
Contracts (IT)   & 2.37     & 53.4\%     & 1.52     & 68.4\%      & 1.01     & 77.8\%     & 0.86      & 81.7\%     & 1.38            & 71.5\%           & 1.14           & 75.9\%            \\
Contracts (PL)   & 2.66      & 50.2\%     & 1.74      & 65.1\%      & 1.17     & 75.0\%     & 0.98      & 79.4\%     & 1.14           & 75.3\%           & 0.94           & 79.5\%            \\
Contracts (PT)   & 2.62     & 49.8\%     & 1.69     & 65.4\%      & 1.14      & 75.4\%     & 0.96      & 79.6\%     & 1.67           & 66.8\%           & 1.46           & 70.2\%            \\
Contracts (RU)   & 1.98     & 58.2\%     & 1.29     & 71.6\%      & 0.85      & 80.5\%     & 0.70      & 84.6\%     & 1.18           & 75.4\%           & 0.99           & 78.8\%            \\
\midrule
Contracts (All)   & 2.26     & 56.6\%     & 1.43     & 71.0\%      & 0.97     & 79.5\%     & 0.87       & 82.9\%     & 1.41          & 71.7\%           & 1.16           & 76.1\%            \\
\midrule
Overall (EN)    & 2.28     & 55.5\%     & 1.44     & 69.9\%      & 0.93     & 79.5\%     & 0.76      & 83.8\%     & 1.32           & 71.9\%           & 1.02           & 77.3\%            \\
Overall (EL)    & 2.02     & 59.4\%     & 1.33     & 71.7\%      & 0.92     & 79.5\%     & 0.81      & 82.9\%     & 1.04           & 77.5\%           & 0.81           & 82.1\%            \\
Overall (DE)    & 2.53     & 52.1\%     & 1.60     & 67.5\%      & 1.04     & 77.5\%     & 0.87      & 81.7\%     & 1.27           & 73.4\%           & 0.99           & 78.6\%            \\
Overall (FR)    & 1.92     & 61.6\%     & 1.20     & 74.6\%      & 0.78     & 82.6\%     & 0.69      & 85.7\%     & 1.09           & 76.8\%           & 0.86            & 81.2\%            \\
Overall (ES)    & 2.12     & 57.9\%     & 1.36     & 71.2\%      & 0.92     & 79.6\%     & 0.81      & 83.0\%     & 1.28           & 73.1\%           & 1.03            & 77.6\%            \\
Overall (NL)    & 2.52     & 52.4\%     & 1.60     & 67.4\%      & 1.06     & 77.1\%     & 0.91      & 81.1\%     & 1.37           & 72.0\%           & 1.06           & 77.4\%            \\
Overall (IT)    & 2.20    & 56.6\%     & 1.40     & 70.7\%      & 0.94     & 79.3\%     & 0.83      & 82.8\%     & 1.30           & 72.8\%           & 1.03           & 77.8\%            \\
Overall (PL)    & 2.41     & 54.3\%     & 1.56     & 68.3\%      & 1.04     & 77.6\%     & 0.86      & 81.7\%     & 1.15           & 75.9\%           & 0.90           & 80.5\%            \\
Overall (PT)    & 2.37     & 54.4\%     & 1.53     & 68.8\%      & 1.05     & 77.7\%     & 0.93      & 81.2\%     & 1.47           & 69.9\%           & 1.22           & 74.4\%            \\
Overall (RU)    & 1.98     & 58.2\%     & 1.29     & 71.6\%      & 0.85      & 80.5\%     & 0.70      & 84.6\%     & 1.18           & 75.4\%           & 0.99           & 78.8\%            \\ 
\midrule
Overall         & 2.37     & 54.9\%     & 1.53     & 69.0\%      & 1.03     & 77.9\%     & 0.90      & 81.5\%     & 1.23           & 74.0\%           & 0.96           & 79.0\%            \\
\bottomrule
\end{tabular}
}
\caption{Masked-Language-Models Validation Performance Scores (Cross-Entropy Loss, Accuracy).}
\label{tab:detailed_mlm}
\end{table*}

\end{document}